\documentclass[11pt,a4paper]{binb}

\usepackage{amssymb}
\usepackage{bm}
\usepackage{nicefrac}
\usepackage{multirow}
\usepackage{wrapfig}
\usepackage{fvextra}
\usepackage{bbm}
\usepackage{cleveref}
\usepackage{multicol}

\definecolor{PromptOrange}{HTML}{F26A00}
\definecolor{PromptBg}{HTML}{FFF3DE}
\definecolor{SectionBg}{HTML}{FFF8F2}
\definecolor{beijingblue}{RGB}{220,235,247}

\newcommand{\pmark}{\textcolor{orange!85!black}{\ding{108}}}
\newcommand{\singleagenticon}{\raisebox{-0.25ex}{\includegraphics[height=1.35em]{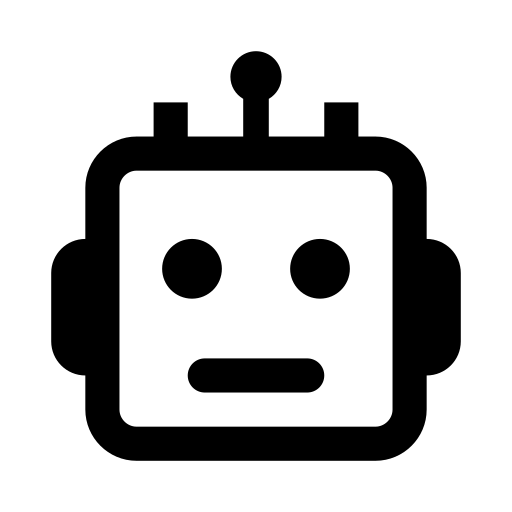}}}
\newcommand{\nus}{\raisebox{-0.25ex}{\includegraphics[height=0.85em]{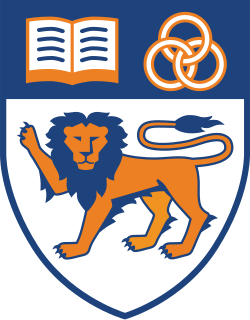}}}
\newcommand{\ntu}{\raisebox{-0.25ex}{\includegraphics[height=0.85em]{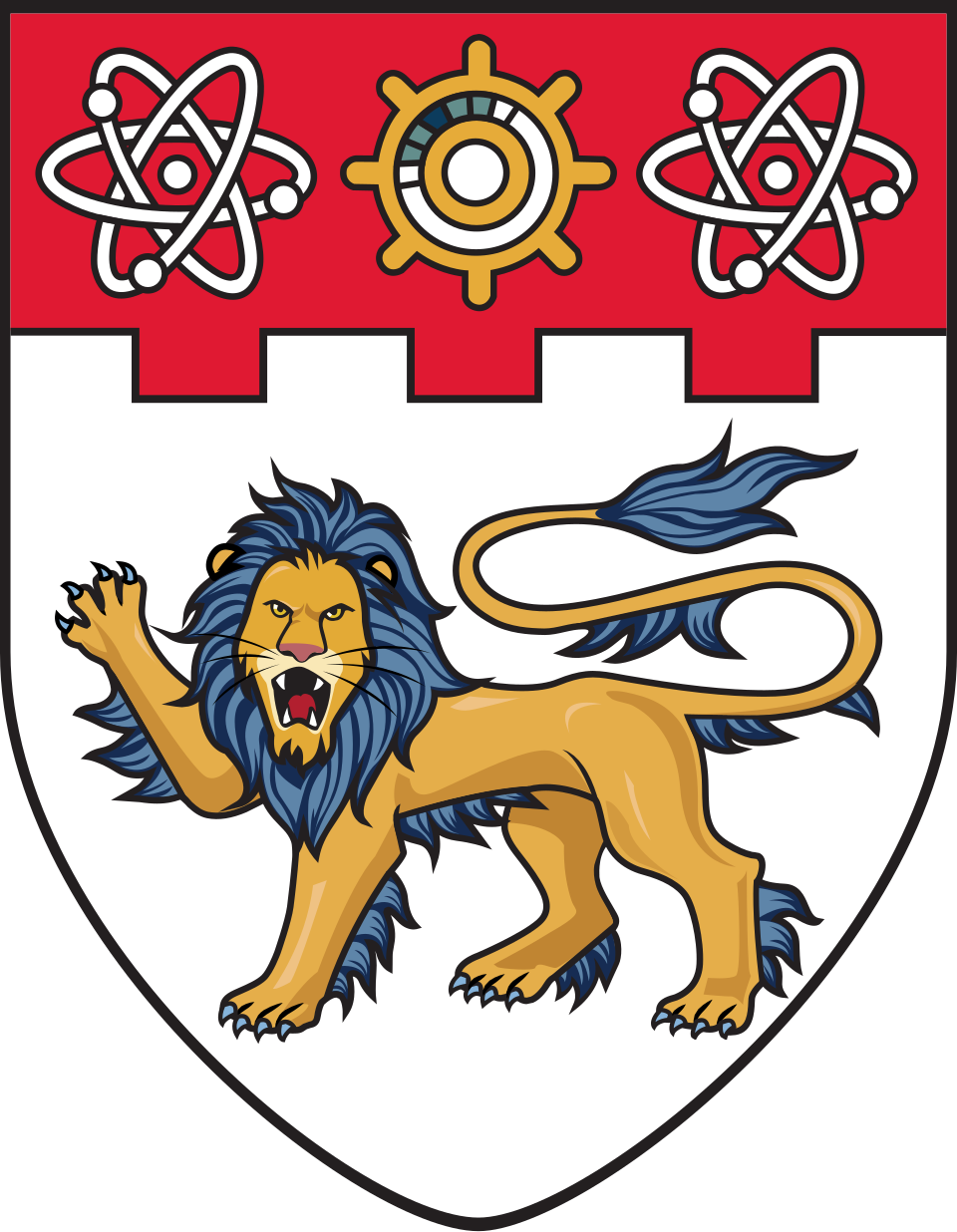}}}
\newcommand{\multiagenticon}{\singleagenticon\hspace{-0.35em}\singleagenticon}

\DefineVerbatimEnvironment{PromptVerbatim}{Verbatim}{
  fontsize=\footnotesize,
  breaklines=true,
  breaksymbolleft={},
  breaksymbolright={}
}

\newcommand{\lmtt}[1]{\texttt{#1}}
\newcommand{\ourmethod}{\lmtt{Mem-W}\xspace}
\newtcolorbox{researchquestionbox}{
  colback=blue!3,
  colframe=blue!18,
  boxrule=0.5pt,
  arc=2pt,
  left=3pt,
  right=3pt,
  top=3pt,
  bottom=3pt,
  width=0.94\linewidth,
}

\newcommand{\githubicon}{\raisebox{-1.5pt}{\includegraphics[height=1.05em]{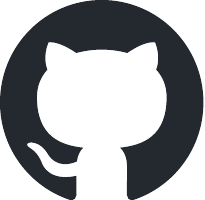}}}
\newcommand{\hficon}{\raisebox{-1.5pt}{\includegraphics[height=1.05em]{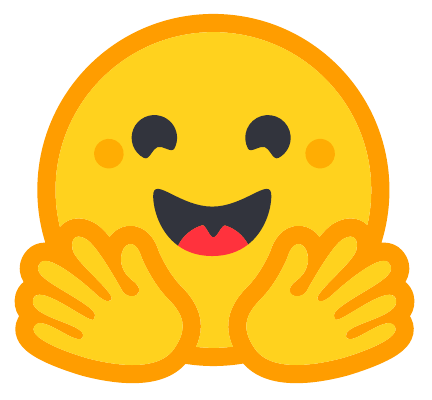}}}
\newcommand{\projectlinks}{%
  \begin{center}
  \begin{tabular}{ccc}
  \href{\ghlink}{\githubicon~GitHub} &
  \href{\hflink}{\hficon~Mem-W-4B} &
  \href{\hflinkt}{\hficon~Mem-W-8B}
  \end{tabular}
  \end{center}%
}

\newcommand{\ghlink}{https://github.com/bingreeky/Mem-W}
\newcommand{\hflinkt}{https://huggingface.co/lunovelle/Mem-W-8B}
\newcommand{\hflink}{https://huggingface.co/lunovelle/Mem-W-4B}

\newcommand{\toolicon}[1]{\raisebox{-0.2ex}{\includegraphics[height=1.8ex]{fig/#1.png}}}
\newcommand{\qwenmodel}{\toolicon{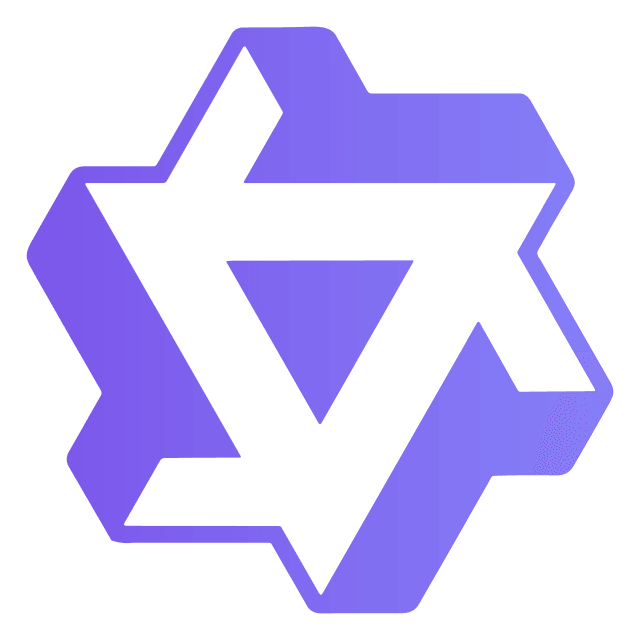}}

\newcommand{\venusmodel}{\toolicon{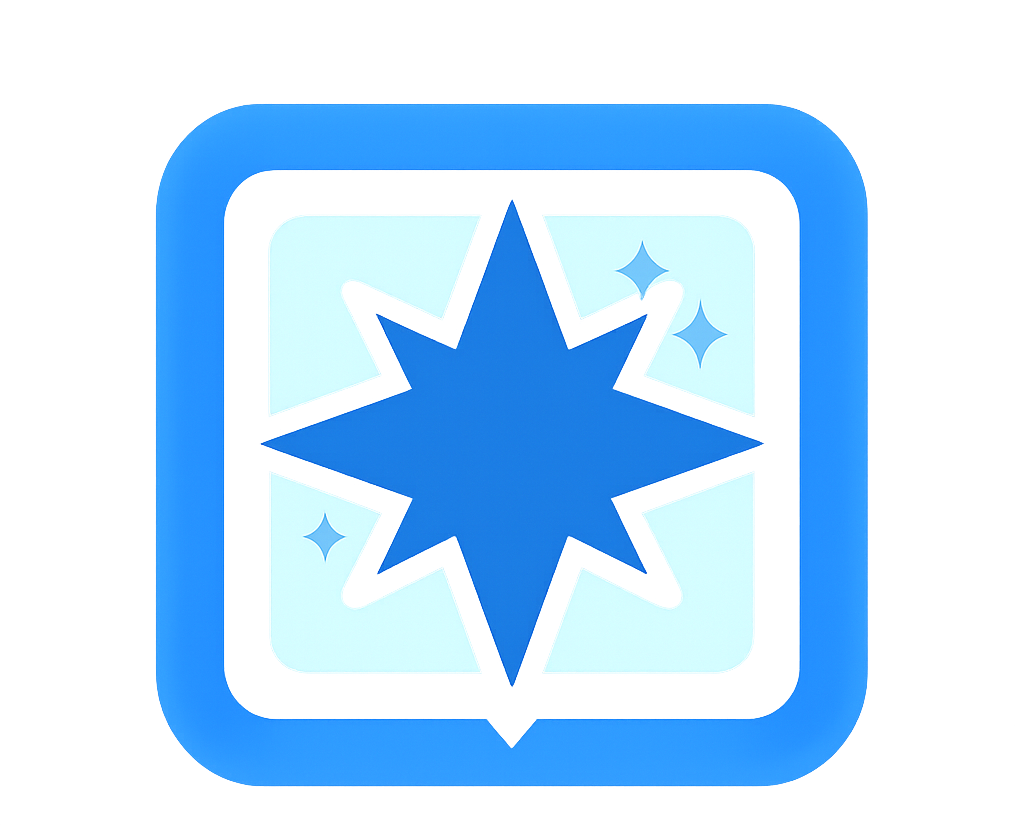}}

\newcommand{\PlaceFirstPageLogo}{%
  \AddToShipoutPictureFG*{%
    \put(\LenToUnit{\dimexpr1in+\oddsidemargin\relax},
         \LenToUnit{\dimexpr\paperheight-1.2cm\relax}){%
      \makebox(0,0)[lt]{\includegraphics[width=2.8cm]{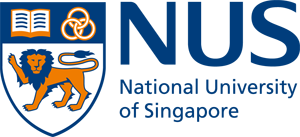}}%
    }%
    \put(\LenToUnit{\dimexpr1in+\oddsidemargin+\textwidth\relax},
         \LenToUnit{\dimexpr\paperheight-2.2cm\relax}){%
      \makebox(0,0)[rt]{\small 2026-5}%
    }%
    \put(\LenToUnit{\dimexpr1in+\oddsidemargin\relax},
         \LenToUnit{\dimexpr\paperheight-2.5cm\relax}){%
      \makebox(0,0)[lt]{\rule{\textwidth}{0.4pt}}%
    }%
  }%
}

\title{\textbf{Mem-W: Latent Memory-Native GUI Agents}}
\author{\nus~{LV-NUS Lab}, NUS \& \ntu~NTU}
\date{\vspace{-3em}}

\begin{document}

\PlaceFirstPageLogo

\maketitle
\vspace{-0.4em}

\projectlinks

\vspace{1em}
\begin{abstract}
\begin{tcolorbox}[colback=tealblueLight,boxrule=0pt,frame hidden,sharp corners,left=1em,right=1em,top=0.8em,bottom=0.8em]
GUI agents are beginning to operate the web, mobile, and desktop as interactive worlds, where successful control depends on carrying forward visual, procedural, and task-level evidence beyond the fleeting present screen.
Yet most agents still treat memory as an external, human-readable artifact: histories are summarized, categorized, retrieved, and reinserted as text or structured records before being encoded again by the policy.
This creates a mismatch between the representational form in which experience is stored and the latent embedding sequence over which modern GUI policies actually act.
We introduce \ourmethod, a series of latent-memory-native GUI agents that treat memory as part of the agent's continuous context rather than as an auxiliary symbolic scaffold.
\ourmethod weaves both historical trajectories (as experiential memory) and in-session segments (as working memory) into compact memory tokens through a shared trajectory-to-latent compressor.
These tokens are woven with the current GUI observation and local context into one continuous embedding sequence, allowing the agent to read successes, failures, and unfinished progress through the same machine-native interface. \ourmethod is trained with self-distillation and outcome-aware supervision to preserve decision-relevant state while filtering memory toward evidence that truly supports task success.
Across four web and mobile navigation benchmarks, \ourmethod consistently improves diverse backbones and memory-enhanced baselines, with gains of up to $+30.0$, suggesting that latent-context-native memory can serve as a scalable foundation for long-horizon GUI agency.
\end{tcolorbox}
\end{abstract}
\vspace{+1.5em}

\section{Introduction}
\vspace{-0.4em}
GUI agents have begun to turn vision-language models (VLMs) into operators of web,
mobile, and desktop interfaces~\cite{nguyen-etal-2025-gui}. Yet long-horizon GUI control is increasingly a context-interface problem, not only a perception or planning problem on a single screen.
A task may require remembering constraints stated several screens ago, preserving partial progress after navigation~\citep{huang2025guikvefficientguiagents,zhou2026efficientlonghorizonguiagents}, reusing a procedure observed
in a similar application, or avoiding an action pattern that previously led to
failure~\citep{cheng2026mgamemorydrivenguiagent,zhu2026hybridselfevolvingstructuredmemory}. The current observation is therefore insufficient, while the full raw
history is too long, noisy, and visually redundant to be carried indefinitely.
The central question is how past interaction evidence should be represented so that it becomes part of the agent's operational context, rather than a detached record to be repeatedly translated back into model inputs.

Recent work has advanced agent memory along two main axes.
One line maintains short-horizon state through explicit, human-readable structures over recent interaction, including graphs, lists, or layered records such as a \textit{task-progress layer}, a \textit{task-status layer}, and a \textit{vision layer}~\citep{wang2025historyawarereasoningguiagents,shi2026androtmeminteractiontrajectoriesanchored,li2026echotrailguibuildingactionablememory,li2026deepagentgeneralreasoningagent,yuan2025memsearchertrainingllmsreason}.
Another line retrieves historical trajectories as long-horizon procedural, experiential memory, often organizing them into reusable abstractions such as a \textit{tool-use experience layer}, \textit{failure reminder layer}, or a \textit{high-level planning layer}~\citep{wu2025autoscalingcontinuousmemorygui,fang2026mempexploringagentprocedural}.
These contributions have made memory increasingly effective, yet they retain a deeper architectural commitment: before learning begins, the designer must decide what memory categories exist, how each is represented, and how each should interface with the policy.
As a result, modern GUI memory designs juggle memory state trackers~\citep{zhong2026actionenginereactiveprogrammaticgui,cheng2026mgamemorydrivenguiagent}, summarizers~\citep{gao2025chainofmemoryenhancingguiagents,ouyang2026reasoningbankscalingagentselfevolving,wang2024agentworkflowmemory}, and retrieval indices~\citep{liu2026omnisimplememautoresearchguideddiscoverylifelong,liu2025memversemultimodalmemorylifelong}, each governed by its own format and update rule.
This fixes the agent to a prescribed memory ontology that may not coincide with the structure demanded by downstream tasks~\citep{zhang2025memevolvemetaevolutionagentmemory,pan2026mstar}.
The memory hierarchies that appear natural to the engineer need not be the ones most useful to the agent policy.

This observation suggests a more basic question: instead of designing
separate mechanisms for each memory category, can a GUI agent become latent-context-native, storing and consuming experience in the same continuous space where its policy already operates?
This work posits that \textbf{latent embedding space} offers a natural answer~\cite{zhu2025surveylatentreasoning,chen2025reasoninglanguagecomprehensivesurvey,yu2026latentspacefoundationevolution}. Because the agent already reasons over continuous embeddings, projecting trajectory evidence into the same space eliminates format mismatch between memory and agent policy, while letting end-to-end gradients (rather than hand-written rules) determine what is retained and how memories compose. Under this formulation, working memory and procedural memory differ only in \textit{provenance} rather than \textit{representation}: one is produced online from the current episode prefix, the other retrieved from an external bank of completed trajectories. Both enter the policy as tokens of the same type, and the structure of the resulting memory organization need not be prescribed but can emerge from training. To put the research question more formally:

\begin{center}
\begin{researchquestionbox}
\centering
\emph{Can a unified latent memory space, learned end-to-end, replace
prescribed and heavily manual-engineered memory hierarchies for GUI agents?}
\end{researchquestionbox}
\end{center}

To answer this question, we propose \ourmethod, a series of latent-memory-native, or more broadly latent-context-native, GUI agents that weave
dual-scale GUI evidence directly into the policy context. \ourmethod learns a
trajectory-to-latent compressor that maps variable-length observation-action
segments into fixed-size latent blocks. At inference time, the same
compressor receives both retrieved historical trajectories and the growing
prefix of the ongoing episode, projecting each into token-efficient latent
representations. The former freely becomes whatever \textbf{\ding{95} experiential signal} the
policy requires (\textit{e.g.}, procedures to imitate, failures to avoid); the latter distills \textbf{\ding{96} working memory} state
without exploding the context window. A shared latent space thus unifies
\textbf{dual-scale GUI memory} (\ding{95} + \ding{96}) with no hand-engineered categorization or
role-specific modules. The training of \ourmethod is driven by two complementary signals:
\textit{self-distillation} from richer raw contexts and \textit{outcome-aware supervision}
that encodes procedural polarity between successful and failed evidence. In this way, memory is no longer an external scaffold imposed on the agent, but a latent context substrate through which experience, state, and action are jointly organized for GUI control.
Our contributions can be summarized as follows:

\vspace{-0.5em}
\begin{itemize}[leftmargin=*,itemsep=-0.1em]
  \item \textbf{Unified formulation.} We recast the increasingly intricate hierarchy of GUI memory as a single latent representation problem, arguing that short-horizon working memory and long-horizon experiential memory can be compactly grounded in one learnable space, where task-relevant structure emerges without prescribed taxonomies.

  \item \textbf{Learnable instantiation.} We instantiate this formulation with \ourmethod, where a shared trajectory-to-latent compressor projects both retrieved historical trajectories and online episode prefixes into a unified latent space, trained through self-distillation and outcome-aware supervision.

  \item \textbf{Empirical evidence.} Experiments across four web and mobile navigation benchmarks show consistent gains of up to $+30.0$ over mainstream GUI agents and hand-engineered memory architectures, while further demonstrating favorable scaling behavior as the experiential memory bank grows with well-controlled inference delay.
\end{itemize}
\section{Related Work}
\vspace{-0.4em}
\paragraph{GUI Agent Memory.}
Drawing on existing memory taxonomies~\citep{hu2026memoryageaiagents,wu2025humanmemoryaimemory}, we group memory architectures for GUI agents into two broad families.
\textbf{(I)~Working memory} mitigates the token growth of long visual interaction histories within a single episode: some methods prune or crop historical screenshots at the visual-token level~\citep{xu2026spatiotemporaltokenpruningefficient,song2026ccpo}, others compress history through KV-cache mechanisms~\citep{huang2025guikvefficientguiagents,zhou2026efficientlonghorizonguiagents}, while AndroTMem~\citep{shi2026androtmeminteractiontrajectoriesanchored} preserves causally linked state anchors and MGA~\citep{cheng2026mgamemorydrivenguiagent} stores structured screenshot--spatial--memory tuples.
\textbf{(II)~Experiential memory} supports cross-session reuse of past experience, but appears in heterogeneous forms: AWM~\citep{wang2024agentworkflowmemory}, ReasoningBank~\citep{ouyang2026reasoningbankscalingagentselfevolving}, ExpeL~\citep{zhao2024expelllmagentsexperiential}, Memp~\citep{fang2026mempexploringagentprocedural}, and EchoTrail-GUI~\citep{li2026echotrailguibuildingactionablememory} distill reusable workflows, lessons, or scripts; and HyMEM~\citep{zhu2026hybrid} combines symbolic strategies with continuous trajectory embeddings.
Despite these advances, existing systems still prescribe, before learning, what memory types exist, how they are represented, and how they are filtered or composed---a design choice that has been shown not always to align with the memory structures demanded by downstream tasks~\citep{zhang2025memevolvemetaevolutionagentmemory,pan2026mstar}.
\ourmethod departs from this ontology-first paradigm by projecting all memory evidence into a shared learnable latent space, allowing task-relevant memory organization to emerge from end-to-end optimization.

\begin{table}[!h]
\centering
\scriptsize
\setlength{\tabcolsep}{3.2pt}
\renewcommand{\arraystretch}{1.12}
\caption{\small Comparison of representative memory mechanisms for GUI and multimodal agents. Working memory denotes in-session state retention, experiential memory denotes cross-session or retrieved task experience, latent memory denotes continuous memory tokens or embeddings, learnable indicates that the memory representation itself is trained rather than shaped only by predefined rules, and multimodal indicates support for GUI or other visual memmory. \cmark, \xmark, and \pmark denote full, absent, and partial support.}
\label{tab:related_memory_comparison}
\begin{adjustbox}{max width=\textwidth}
\begin{tabular}{lcccccccl}
\toprule
\textbf{Method} & \textbf{Working} & \textbf{Exper.} & \textbf{Latent} & \textbf{Native ctx.} & \textbf{Learnable} & \textbf{Multimodal} & \textbf{Agent} & \textbf{Memory interface} \\
\midrule
AWM~\citep{wang2024agentworkflowmemory}
& \xmark & \cmark & \xmark & \xmark & \xmark & \pmark & \singleagenticon & Reusable textual workflows \\
ReasoningBank~\citep{ouyang2026reasoningbankscalingagentselfevolving}
& \xmark & \cmark & \xmark & \xmark & \xmark & \xmark & \singleagenticon & Reasoning memories from self-evaluated experience \\
Memp~\citep{fang2026mempexploringagentprocedural}
& \xmark & \cmark & \xmark & \xmark & \xmark & \pmark & \singleagenticon & Procedural instructions and script abstractions \\
EchoTrail-GUI~\citep{li2026echotrailguibuildingactionablememory}
& \xmark & \cmark & \xmark & \xmark & \xmark & \cmark & \singleagenticon & Reward-validated actionable trajectory memory \\
AndroTMem~\citep{shi2026androtmeminteractiontrajectoriesanchored}
& \cmark & \xmark & \xmark & \xmark & \xmark & \cmark & \singleagenticon & Causally linked state anchors \\
MGA~\citep{cheng2026mgamemorydrivenguiagent}
& \cmark & \xmark & \xmark & \xmark & \xmark & \cmark & \singleagenticon & Dynamic structured state memory \\
HyMEM~\citep{zhu2026hybrid}
& \cmark & \cmark & \pmark & \xmark & \pmark & \cmark & \singleagenticon & Graph memory with trajectory embeddings \\
CoMEM~\citep{wu2025autoscalingcontinuousmemorygui,wu2025generalcontinuousmemoryvisionlanguage}
& \xmark & \cmark & \cmark & \cmark & \cmark & \cmark & \singleagenticon & Retrieved continuous trajectory embeddings \\
VisMem~\citep{yu2026vismemlatentvisionmemory}
& \cmark & \pmark & \cmark & \cmark & \cmark & \cmark & \singleagenticon & Short-term and long-term latent vision memory \\
L$^{2}$-VMAS~\citep{yu2026duallatentmemoryvisual}
& \pmark & \xmark & \cmark & \cmark & \cmark & \cmark & \multiagenticon & Dual latent memories for visual multi-agent communication \\
\textbf{\ourmethod}
& \cmark & \cmark & \cmark & \cmark & \cmark & \cmark & \singleagenticon & Unified latent context tokens \\
\bottomrule
\end{tabular}
\end{adjustbox}
\vspace{-0.4em}
\end{table}

\vspace{-0.2em}
\paragraph{Latent memory.}
Latent space has emerged as a promising memory
substrate~\citep{hu2026memoryageaiagents,yu2026latentspacefoundationevolution},
owing to its token efficiency, machine-native format, and end-to-end
learnability. For textual agents, a growing body of work encodes
cross-session experience as latent
embeddings, including NextMem~\citep{zhang2026nextmemlatentfactualmemory}, MemGen~\citep{zhang2026memgen}, SoftCoT/SoftCoT++~\citep{xu2025softcotsoftchainofthoughtefficient,xu2025softcottesttimescalingsoft}, FlashMem~\citep{hou2026flashmemdistillingintrinsiclatent} and others~\citep{xu2026gmemllmgatedlatentmemory,wang2025mextendingmemoryllmscalable,wang2025selfupdatablelargelanguagemodels}.
The idea extends naturally to multimodal agents:
L2-VMAS~\citep{yu2026duallatentmemoryvisual} and
VisMem~\citep{yu2026vismemlatentvisionmemory} encode visual cues and
trajectories through latent tokens, while
CoMEM~\citep{wu2025autoscalingcontinuousmemorygui,wu2025generalcontinuousmemoryvisionlanguage}
compresses historical trajectories into continuous embeddings to serve as
experiential memory. Our work draws inspiration from this line, especially CoMEM~\citep{wu2025autoscalingcontinuousmemorygui}. We also benefit from the scaled GUI trajectory resources released by this work, which support part of our training pipeline as detailed later. However, \ourmethod differs fundamentally from prior work: whereas existing methods typically employ latent representations for a single memory role, \ourmethod uses a shared compressor and a unified latent space to subsume both working memory and experiential memory, allowing task-relevant memory structure to emerge through end-to-end training. \Cref{tab:related_memory_comparison} summarizes the comparison between \ourmethod and contemporary memory-based agent methods.

\begin{figure}[!t]
    \centering
    \includegraphics[width=1\linewidth]{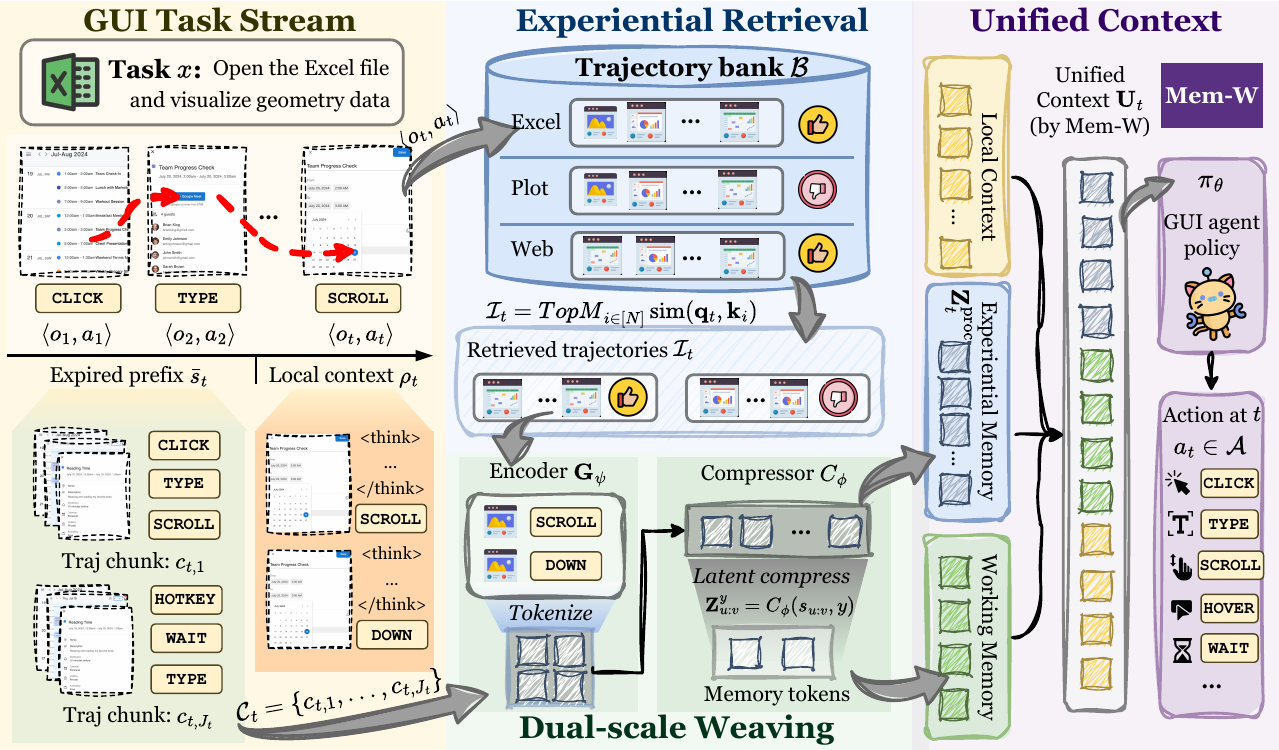}
    \caption{\small Overview of \ourmethod. Given a GUI task stream, \ourmethod retrieves relevant historical trajectories $\mathcal{B}$ as experiential memory and compresses expired in-session segments $\bar{s}_t$ as working memory through compressor $C_\phi$. The resulting memory tokens are woven with the local context into a unified embedding sequence, enabling the GUI policy to act over both prior experience and current progress.}
    \vspace{-0.2em}
    \label{fig:framework}
\end{figure}

\vspace{-0.4em}
\section{Method}
\vspace{-0.5em}
As illustrated in \Cref{fig:framework}, \ourmethod instantiates a latent-context-native interface for GUI agents.
The goal is not to add another human-readable memory store beside the policy, but to convert interaction evidence itself into soft context tokens that live in the agent's input embedding space.
Short-horizon episode history and long-horizon retrieved experience are therefore exposed to the policy through the same continuous interface, allowing the agent to attend to them as native context rather than as externally serialized notes.
This section formalizes necessary notations ($\triangleright$ \Cref{sec:prelim}) and then presents the compressor ($\triangleright$ \Cref{sec:method-compressor}), inference procedure ($\triangleright$ \Cref{sec:method-inference}), and training recipe ($\triangleright$ \Cref{sec:method-training}).

\vspace{-0.4em}
\subsection{Preliminaries and Overview}\label{sec:prelim}
\vspace{-0.4em}

We consider an interactive GUI task specified by an instruction $x$. At step
$t$, the agent receives an observation $o_t\in\mathcal{O}$, such as a screenshot,
and predicts an executable action $a_t\in\mathcal{A}$. We write one interaction
event as $e_t=(o_t,a_t)$ and a contiguous event sequence as
$s_{u:v}=(e_u,\ldots,e_v)$. A completed trajectory is denoted by
$\tau=(x,s_{1:T},y)$, where
$y\in\mathcal{Y}=\{\mathrm{succ},\mathrm{fail}\}$ is the terminal outcome label.
The current episode is unlabeled before termination, while trajectories stored
in an offline memory bank carry success or failure labels.

Let $\bm{\Pi}_\theta$ denote the frozen GUI agent. We use
$\mathbf{G}_\theta$ for its visual--textual encoder and
$\mathbf{E}_\theta(x,o_t,\rho_t)$ for its ordinary input embeddings, where
$\rho_t$ is the bounded raw context retained around the current step. The goal
of \ourmethod is to equip this frozen agent with a compact latent memory
interface, without updating the agent parameters $\theta$.
Formally:
\begin{equation}
  \mathfrak{S}
  =
  \bigcup_{\ell\geq 1}(\mathcal{O}\times\mathcal{A})^\ell,
  \qquad
  C_\phi:\mathfrak{S}\times(\mathcal{Y}\cup\{\varnothing\})
  \rightarrow
  \mathbb{R}^{K\times d},
  \label{eq:memw-compressor-abstract}
\end{equation}
where $\mathfrak{S}$ is the space of finite GUI event sequences, $K$ is the
per-segment latent-token budget, $d$ is the embedding dimension of the GUI
agent, and $\varnothing$ marks a segment whose final outcome is unknown.
Action serialization $\alpha(\cdot)$ and outcome embedding $\gamma(\cdot)$ are
implemented inside the compressor and are omitted from the abstract signature
for notational compactness.
The compressor is not a natural-language summarizer. It maps procedural GUI
evidence into soft tokens that can be directly consumed by the frozen agent.
At each decision step, \ourmethod concatenates two sources of latent memory:
\begin{equation}
  \mathcal{M}_t
  =
  \bigl[
  \mathcal{M}^{\mathrm{proc}}_t
  \;;\;
  \mathcal{M}^{\mathrm{work}}_t
  \bigr],
  \label{eq:dual-scale-memory-space}
\end{equation}
where $\mathcal{M}^{\mathrm{proc}}_t$ is distilled from retrieved historical
trajectories and $\mathcal{M}^{\mathrm{work}}_t$ is distilled from the expired
prefix of the current episode. The two sources differ by provenance and role,
not by representation: every individual $K\times d$ block in either memory is
an output of the same compressor $C_\phi$.
Training therefore focuses on the compressor parameters $\phi$, while keeping
the GUI agent $\bm{\Pi}_\theta$ frozen. The following subsections instantiate
this interface with a query-based trajectory compressor, describe how procedural
and working memories are woven into the agent context, and then give the
two-stage training procedure.

\vspace{-0.3em}
\subsection{Unified Trajectory-to-Latent Compressor}
\label{sec:method-compressor}
\vspace{-0.3em}

The compressor fixes the latent budget of each selected trajectory segment and
places both memory scales in a shared continuous token space. Given
$s_{u:v}=(e_u,\ldots,e_v)$ with $e_i=(o_i,a_i)$, we pair each observation with a
structured action serialization $\alpha(a_i)$ that records the action type and
arguments. We also attach an outcome embedding
$\gamma(y)\in\mathbb{R}^{h}$ for
$y\in\mathcal{Y}\cup\{\varnothing\}$. The unknown marker $\varnothing$ is used
for online working-memory segments whose terminal outcome is not yet available.
The frozen encoder $\mathbf{G}_\theta$ converts the observation--action segment
into contextual features:
\begin{equation}
  \mathbf{H}_{u:v}^{y}
  =
  \mathbf{G}_\theta
  \bigl(
  (o_u,\alpha(a_u)),\ldots,(o_v,\alpha(a_v));\,\gamma(y)
  \bigr)
  \in \mathbb{R}^{N_{u:v}\times h},
  \label{eq:event-stream-encoding}
\end{equation}
where $N_{u:v}$ is the number of encoder features and $h$ is the hidden
dimension. The outcome embedding allows completed successful and failed
trajectories to be encoded differently, while preserving the same interface for
unlabeled online segments.
We instantiate $C_\phi$ with a lightweight query transformer following the
Q-Former design~\citep{li2023blip2bootstrappinglanguageimagepretraining}. With
$K$ learned queries $\mathbf{Q}\in\mathbb{R}^{K\times h}$, the compressor
extracts a fixed-size latent block and projects it into the agent embedding
space:
\begin{equation}
  \mathbf{Z}_{u:v}^{y}
  =
  C_\phi(s_{u:v},y)
  =
  \mathbf{P}_\phi
  \left(
  \operatorname{QFormer}_\phi
  \left(
  \mathbf{Q}+\mathbf{1}_K\gamma(y)^\top,\;
  \mathbf{H}_{u:v}^{y}
  \right)
  \right)
  \in \mathbb{R}^{K\times d},
  \label{eq:qformer-compressor}
\end{equation}
where $\mathbf{P}_\phi:\mathbb{R}^{h}\rightarrow\mathbb{R}^{d}$ is applied
token-wise and $\mathbf{1}_K\in\mathbb{R}^{K\times 1}$ broadcasts the outcome
embedding to all query slots. The output
$\mathbf{Z}_{u:v}^{y}=[\mathbf{z}_1;\ldots;\mathbf{z}_K]$ is never decoded into
text; it is already a block of soft memory tokens for the agent.

\vspace{-0.3em}
\subsection{Dual-Scale Memory Weaving at Inference}
\label{sec:method-inference}
\vspace{-0.3em}

Given the compressor $C_\phi$, inference consists of selecting a small number
of relevant segments, compressing them into latent blocks, and placing those
blocks before the ordinary agent context.

\vspace{-0.4em}
\paragraph{Procedural memory.}
The procedural memory is retrieved from an external trajectory bank
$\mathcal{B}=\{(x_i,s^i_{1:T_i},y_i)\}_{i=1}^{N}$. We reuse the frozen encoder
$\mathbf{G}_\theta$ to form retrieval keys and the current query:
\begin{equation}
  \mathbf{k}_i
  =
  \operatorname{pool}
  \bigl(
  \mathbf{G}_\theta(x_i,s^i_{1:T_i})
  \bigr),\;
  \mathbf{q}_t
  =
  \operatorname{pool}
  \bigl(
  \mathbf{G}_\theta(x,o_t,\rho_t)
  \bigr),\;
  \mathcal{I}_t
  =
  \operatorname*{Top\text{-}M}_{i\in[N]}
  \operatorname{sim}(\mathbf{q}_t,\mathbf{k}_i),
  \label{eq:procedural-retrieval}
\end{equation}
where $M$ is the number of retrieved trajectories and
$\operatorname{pool}(\cdot)$ extracts the last hidden state. The retrieval
encoder is frozen, and the discrete Top-$M$ operation is not optimized by
gradient descent.
Let $(i_1,\ldots,i_M)$ be the retrieved indices sorted by descending similarity.
The procedural memory is obtained by compressing the corresponding trajectories
and concatenating them in retrieval-rank order:
\begin{equation}
  \mathbf{Z}^{\mathrm{proc}}_t
  =
  \bigl[
  C_\phi(s^{i_1}_{1:T_{i_1}},y_{i_1})
  \;;\;\cdots\;;\;
  C_\phi(s^{i_M}_{1:T_{i_M}},y_{i_M})
  \bigr]
  \in \mathbb{R}^{MK\times d}.
  \label{eq:procedural-latents}
\end{equation}
Successful and failed trajectories are therefore represented in the same latent
space, with their outcome information encoded through $\gamma(y)$ rather than
through hand-designed filtering rules.

\vspace{-0.4em}
\paragraph{Working memory.}
The working-memory path uses the same compressor. We keep the most recent $L$
interaction steps in raw form,
\begin{equation}
  \rho_t=s_{\max(1,t-L):t-1},
\end{equation}
and compress only the expired prefix. When $t>L+1$, the expired prefix is
$\bar{s}_t=s_{1:t-L-1}$; otherwise it is empty. We partition $\bar{s}_t$ into
non-overlapping chunks of at most $W$ steps,
$\mathcal{C}_t=\{c_{t,1},\ldots,c_{t,J_t}\}$, where
$J_t=\lceil |\bar{s}_t|/W\rceil$. If $\bar{s}_t$ is empty, then $J_t=0$ and
the working-memory block has zero rows.
The chunks are compressed with the unknown-outcome marker and concatenated in
temporal order:
\begin{equation}
  \mathbf{Z}^{\mathrm{work}}_t
  =
  \bigl[
  C_\phi(c_{t,1},\varnothing)
  \;;\;\cdots\;;\;
  C_\phi(c_{t,J_t},\varnothing)
  \bigr]
  \in \mathbb{R}^{J_tK\times d}.
  \label{eq:working-latents}
\end{equation}
Thus, historical procedural evidence and the agent's own past are compressed by
the same mechanism, while remaining distinguishable by source.

\vspace{-0.4em}
\paragraph{Latent weaving.}
Let
$\mathbf{E}_\theta(x,o_t,\rho_t)\in\mathbb{R}^{n_t\times d}$ be the ordinary
input embeddings of the frozen GUI agent, including the current screenshot and
the retained raw context. We add learned source embeddings
$\mathbf{b}^{\mathrm{proc}},\mathbf{b}^{\mathrm{work}}\in\mathbb{R}^{d}$ and
construct the full latent-augmented input:
\begin{equation}
  \mathbf{U}_t
  =
  \bigl[
  \mathbf{Z}^{\mathrm{proc}}_t
  +
  \mathbf{1}_{MK}(\mathbf{b}^{\mathrm{proc}})^\top
  \;;\;
  \mathbf{Z}^{\mathrm{work}}_t
  +
  \mathbf{1}_{J_tK}(\mathbf{b}^{\mathrm{work}})^\top
  \;;\;
  \mathbf{E}_\theta(x,o_t,\rho_t)
  \bigr],
  \label{eq:weaved-policy}
\end{equation}
where $\mathbf{1}_n\in\mathbb{R}^{n\times 1}$ broadcasts a source embedding to
all rows of the corresponding block. The frozen agent then predicts the next
action by
\begin{equation}
  \pi_{\theta,\phi}(a_t\mid\mathbf{U}_t)
  =
  \bm{\Pi}_\theta(a_t\mid\mathbf{U}_t).
  \label{eq:policy-shorthand}
\end{equation}

Each selected segment contributes exactly $K$ latent tokens, so the total memory
context contains $(M+J_t)K$ latent tokens. In practice, we cap $M+J_t$ to a
fixed constant, ensuring that the aggregate latent overhead remains bounded.

\vspace{-0.3em}
\subsection{Training Latent-Memory GUI Agents}
\label{sec:method-training}
\vspace{-0.3em}

Since $\bm{\Pi}_\theta$ is frozen, training only adjusts the
compressor parameters $\phi$. The objective is to make the latent blocks
decision-relevant and usable by the frozen policy via two-stage training:

\vspace{-0.4em}
\paragraph{Stage 1: Compression fidelity via self-distillation.}
The first stage teaches the latent memory paths to replace raw historical
context. We use the same frozen agent as both teacher and student. The teacher
observes an extended raw window $\chi_t^{\mathrm{raw}}=(x,\,s_{\max(1,t-L'):t-1},\,o_t),\;L'\gg L$,
while the student observes the latent-augmented input $\mathbf{U}_t$ from
\Cref{eq:weaved-policy}, where both working-memory blocks and retrieved
procedural-memory blocks are constructed by $C_\phi$. 
The self-distillation loss is:
\begin{equation}
  \mathcal{L}_{\mathrm{sd}}
  =
  \mathbb{E}_{(\tau,t)}
  \left[
  \ell_{\mathrm{gui}}
  \bigl(
  \bm{\Pi}_\theta(\cdot\mid\mathbf{U}_t),\;
  a_t^\star
  \bigr)
  +
  \lambda
  D_{\mathrm{KL}}
  \left(
  \operatorname{sg}
  \bigl[
  \bm{\Pi}_\theta(\cdot\mid\chi_t^{\mathrm{raw}})
  \bigr]
  \;\Vert\;
  \bm{\Pi}_\theta(\cdot\mid\mathbf{U}_t)
  \right)
  \right],
  \label{eq:self-distillation}
\end{equation}
where $\operatorname{sg}[\cdot]$ stops gradients through the teacher path.
Because $\theta$ is frozen, gradients from both the action loss and the KL term
flow only into $\phi$ through the latent tokens in $\mathbf{U}_t$. The
action-level loss anchors the compressor to the target behavior, while the KL
term transfers the teacher's extended-context distribution into the compressed
representation, including both in-session working state and the retrieved
experiential evidence.

\vspace{-0.4em}
\paragraph{Stage 2: Outcome-aware compressor optimization.}
The second stage also keeps both procedural and working memory active, and further
optimizes their latent encoding with environment feedback. The frozen agent,
conditioned on $\mathbf{U}_t$, rolls out full episodes in the GUI environment
and receives a terminal reward $R(\tau)\in\{0,1\}$. We optimize the compressor
with a policy-gradient objective:
\begin{equation}
  \min_{\phi}
  \;
  -
  \mathbb{E}_{\tau\sim\pi_{\theta,\phi}}
  \left[
  \sum_{t=1}^{T}
  \left(
  \log
  \bm{\Pi}_\theta(a_t\mid\mathbf{U}_t)
  \cdot
  \hat{R}(\tau)
  -
  \beta
  D_{\mathrm{KL}}
  \left(
  \bm{\Pi}_\theta(\cdot\mid\mathbf{U}_t)
  \;\Vert\;
  \pi_{\mathrm{ref}}(\cdot\mid\mathbf{U}^{\mathrm{ref}}_t)
  \right)
  \right)
  \right],
  \label{eq:outcome-aware-rl}
\end{equation}
where $\hat{R}(\tau)=R(\tau)-\bar{R}$ is a baselined reward, $\beta$ controls
the KL regularization strength, and $\pi_{\mathrm{ref}}$ is the frozen
stage-one reference policy. The reference context
$\mathbf{U}^{\mathrm{ref}}_t$ is constructed with the frozen stage-one
compressor, so the KL anchor remains fixed while $C_\phi$ is updated.

This stage does not backpropagate through the environment or the discrete
retrieval operation. Instead, the terminal reward provides a policy-gradient
signal on the sampled action log-probabilities, and gradients reach $\phi$
through the differentiable latent tokens inside $\mathbf{U}_t$. The compressor
therefore learns not only what information to preserve, but also how to encode
success- and failure-conditioned procedural evidence in a way that improves the
frozen agent's task performance.
After training, inference uses the learned compressor $C_\phi$, the frozen
encoder $\mathbf{G}_\theta$ for retrieval, the trajectory bank $\mathcal{B}$,
and the frozen GUI agent $\bm{\Pi}_\theta$. More details on the training process is detailed in \Cref{app:train}.

\begin{table*}[!t]
\centering
\scriptsize
\caption{\small Performance comparison on AndroidControl-v2 (AC-v2) and GUI-Odyssey.
For AC-v2-High/Low, we report Pass@1 and Pass@4, each decomposed into step accuracy (Acc), action type accuracy (Type), and grounding accuracy (Ground). For GUI-Odyssey, we report Type Match (TM) and Exact Match (EM). The baselines marked with $\dagger$ are reused from GUI-Libra.}
\vspace{-0.3em}
\label{tab:ac_high_low_odyssey_merged}
\setlength{\tabcolsep}{2.6pt}
\resizebox{\textwidth}{!}{
\begin{tabular}{l|ccc|ccc|ccc|ccc|cc}
\toprule
\multirow{3}{*}{\textbf{Model}}
& \multicolumn{6}{c|}{\textbf{AC-v2-High}}
& \multicolumn{6}{c|}{\textbf{AC-v2-Low}}
& \multicolumn{2}{c}{\textbf{Odyssey}} \\
\cmidrule(lr){2-7} \cmidrule(lr){8-13} \cmidrule(lr){14-15}
& \multicolumn{3}{c|}{\textbf{Pass@1}}
& \multicolumn{3}{c|}{\textbf{Pass@4}}
& \multicolumn{3}{c|}{\textbf{Pass@1}}
& \multicolumn{3}{c|}{\textbf{Pass@4}}
& \multirow{2}{*}{\textbf{TM}}
& \multirow{2}{*}{\textbf{EM}} \\
\cmidrule(lr){2-4} \cmidrule(lr){5-7} \cmidrule(lr){8-10} \cmidrule(lr){11-13}
& Acc & Type & Ground
& Acc & Type & Ground
& Acc & Type & Ground
& Acc & Type & Ground
&  &  \\
\midrule

\multicolumn{15}{c}{\textbf{Proprietary / Closed-source Models}} \\
\midrule
GPT-4o + UGround$^\dagger$
& 57.00 & -- & -- & 66.30 & -- & --
& 78.40 & -- & -- & 85.40 & -- & --
& -- & - \\

GPT-4.1 + UGround$^\dagger$
& 57.50 & -- & -- & 63.30 & -- & --
& 78.40 & -- & -- & 83.20 & -- & --
& -- & -- \\

GPT-5-mini + UGround$^\dagger$
& 52.80 & -- & -- & 58.80 & -- & --
& 77.10 & -- & -- & 83.20 & -- & --
& -- & -- \\

GPT-5 + UGround$^\dagger$
& 61.30 & -- & -- & 69.40 & -- & --
& 86.20 & -- & -- & 90.00 & -- & --
& -- & -- \\

\midrule
\multicolumn{15}{c}{\textbf{Open-source Models}} \\
\midrule
GUI-R1-3B$^\dagger$
& 40.00 & -- & -- & 54.00 & -- & --
& 55.80 & -- & -- & 71.90 & -- & --
& -- & -- \\

GUI-R1-7B$^\dagger$
& 39.70 & -- & -- & 56.30 & -- & --
& 62.30 & -- & -- & 72.60 & -- & --
& -- & -- \\

Aguvis-7B$^\dagger$
& 37.70 & -- & -- & 43.70 & -- & --
& 48.00 & -- & -- & 48.70 & -- & --
& 26.70 & 13.50 \\

UI-TARS-1.5-7B
& 45.20 & 70.60 & 57.40 & 63.10 & 84.17 & 72.65
& 48.50 & 71.36 & 80.27 & 70.90 & 89.95 & 87.44
& 72.13 & 50.98 \\

GLM-4.1V-9B-Thinking
& 37.20 & 67.84 & 48.43 & 49.00 & 74.87 & 59.64
& 67.10 & 90.20 & 67.26 & 73.40 & 93.22 & 78.48
& 67.26 & 32.88 \\

GUI-Owl-1.5-8B
& 40.45 & 59.05 & 60.09 & 52.76 & 69.60 & 73.09
& 77.14 & 86.68 & 88.34 & 84.42 & 93.47 & 92.38
& 79.24 & 58.95 \\

Qwen3-VL-2B
& 45.48 & 64.07 & 59.64 & 59.80 & 73.62 & 73.99
& 77.64 & 85.18 & 87.00 & 84.92 & 89.95 & 93.72
& 58.85 & 38.70 \\

Qwen3-VL-8B
& 54.80 & 69.35 & 67.71 & 66.10 & 77.89 & 79.82
& 77.60 & 82.91 & 90.27 & 83.20 & 86.93 & 94.17
& 74.96 & 48.73 \\

Qwen3-VL-32B
& 59.05 & 74.62 & 69.96 & 70.10 & 82.16 & 77.58
& 84.42 & 89.95 & 91.83 & 88.69 & 92.21 & 94.17
& 80.62 & 56.69 \\

MAI-UI-2B
& 42.71 & 61.06 & 66.37 & 43.47 & 65.33 & 62.78
& 49.25 & 64.32 & 82.51 & 54.77 & 68.09 & 87.00
& 70.12 & 47.95 \\

Step-GUI-4B
& 59.30 & 74.37 & 69.96 & 71.36 & 83.17 & 80.72
& 68.84 & 75.13 & 88.79 & 82.16 & 86.68 & \textbf{94.62}
& 64.99 & 44.94 \\

GUI-Libra-7B$^\dagger$
& 59.30 & -- & -- & 67.30 & -- & --
& 85.20 & -- & -- & 90.70 & -- & --
& -- & -- \\

GUI-Libra-8B$^\dagger$
& 64.30 & -- & -- & 70.60 & -- & --
& 88.90 & -- & -- & 91.70 & -- & --
& -- & -- \\

UI-Venus-1.5-30B-A3B
& 51.26 & 67.84 & 64.13 & 64.07 & 74.62 & 75.34
& 79.65 & 87.94 & 87.44 & 87.44 & 91.21 & 91.93
& 80.56 & 60.64 \\

\midrule
\multicolumn{15}{c}{\textbf{\ourmethod Models}} \\
\midrule
Qwen3-VL-4B
& 49.30 & 63.57 & 69.06 & 63.30 & 75.88 & 77.13
& 78.90 & 85.93 & 90.58 & 82.40 & 88.69 & 93.72
& 72.89 & 44.73 \\

\rowcolor{LightCyan1}
\textbf{$\hookrightarrow$ \ourmethod-4B (Ours)}
& {63.07} & {74.62} & {70.85}
& {74.87} & {83.92} & {80.72}
& {84.92} & {90.95} & \textbf{92.58}
& {89.20} & {93.72} & {93.72}
& {75.07} & {46.93} \\

UI-Venus-1.5-8B
& 61.06 & 74.62 & 61.06 & 70.10 & 80.40 & 82.51
& 81.66 & 88.44 & 91.03 & 87.19 & 93.22 & 94.17
& 83.12 & 62.45 \\

\rowcolor{LightCyan1}
\textbf{$\hookrightarrow$ \ourmethod-8B (Ours)}
& \textbf{68.59} & \textbf{81.16} & \textbf{71.30}
& \textbf{80.40} & \textbf{88.94} & \textbf{82.96}
& \textbf{87.94} & \textbf{93.72} & {89.24}
& \textbf{94.22} & \textbf{98.24} & \textbf{94.62}
& \textbf{84.88} & \textbf{65.05} \\

\bottomrule
\end{tabular}
}
\end{table*}

\vspace{-0.2em}
\section{Experiments}
\vspace{-0.4em}
\subsection{Experiment Setup}\label{sec:exp-setup}
\vspace{-0.4em}
\paragraph{Training data.}
We train \ourmethod on web trajectories from CoMEM~\citep{wu2025autoscalingcontinuousmemorygui} and the official mobile training set from~\citep{guiodyssey}.
The web split contains $11{,}176$ successful training trajectories across $13$ domains, with a memory bank of $22{,}346$ successful trajectories; the mobile split contains $2{,}489$ successful training trajectories across six task categories, with a memory bank of $4{,}972$ successful trajectories.
Stage~I trains $C_\phi$ for trajectory-to-latent compression via self-distillation, while Stage~II further optimizes $C_\phi$ with outcome-aware supervision.
Further data details are provided in~\Cref{app:training-data}.

\vspace{-0.4em}
\paragraph{Evaluation Benchmark.} Experiments are conducted on four challenging multimodal GUI benchmarks, including two web navigation benchmarks and two mobile navigation benchmarks:
\vspace{-0.3em}
\begin{itemize}[leftmargin=2em,itemsep=-0.1em]
    \item \textbf{Multimodal-Mind2Web}~\citep{deng2023mind2web,zheng2024gpt} is a large-scale benchmark comprising over 2,000 open-ended tasks collected from 137 real-world websites across 31 domains. We directly adopt the test subset used in \citep{wu2025autoscalingcontinuousmemorygui}. The evaluation metric is \textit{success rate}, which measures whether the entire long-horizon task is completed successfully.
    \item \textbf{MMINA}~\citep{tian2025mmina} is designed to evaluate GUI agents on real-world websites. We use its shopping subset for evaluation, with \textit{success rate} as the metric.
    \item \textbf{GUI-Odyssey}~\citep{guiodyssey} evaluates cross-app GUI navigation on mobile devices, covering 6 mobile devices and 212 distinct applications. Its evaluation metrics include \textit{Type Match (TM)}, which assesses whether the predicted action type matches the ground truth, and \textit{Exact Match (EM)}, which further requires all action parameters to be predicted correctly.
    \item \textbf{AndroidControl-v2}~\citep{li2024androidcontrol,yang2026guilibra}, introduced by GUI-Libra~\citep{yang2026guilibra}, revises AndroidControl~\citep{li2024androidcontrol} by correcting label errors and translating non-natural symbolic action histories. It adopts three major metrics, including step-wise accuracy, grounding accuracy and action type accuracy.
\end{itemize}
\vspace{-0.4em}

\paragraph{\ourmethod Details.}
We apply the training procedure described in~\Cref{sec:method-training} to
three mainstream GUI backbones: UI-TARS-1.5-7B~\citep{ui-tars-15-seed},
Qwen3-VL-4B~\citep{Qwen3-VL}, and
UI-Venus-1.5-8B~\citep{gao2026ui}. Notably, the backbone parameters
$\theta$ remain frozen throughout; only the compressor parameters $\phi$
are trained. At evaluation time, all models operate under a ReAct-style
paradigm, outputting structured reasoning followed by a tool invocation
from a predefined GUI action set (see \Cref{app:evaluation-setup}). The latent token count $K$ is set as $8$, the working memory window $L$ equals $3$, chunk size $W$ is $4$. The number of trajectories retrieved ($M$) in  \Cref{eq:procedural-retrieval} is $5$. More  parameters are detailed in \Cref{app:param,app:evaluation-setup}.

\vspace{-0.4em}
\paragraph{Baselines.} We compare \ourmethod with: \textbf{(I) open-source GUI agents}, including UI-TARS-1.5-7B~\citep{ui-tars-15-seed}, Qwen3-VL-2/4/8/32B~\citep{Qwen3-VL}, UI-Venus-1.5-8B/30B-A3B~\citep{gao2026ui}, GLM-4.1V-9B-Thinking~\citep{vteam2025glm45vglm41vthinkingversatilemultimodal}, GUI-R1-3/7B~\citep{luo2025gui-r1}, MAI-UI-2B~\citep{zhou2025maiuitechnicalreportrealworld}, Step-GUI-4B~\citep{yan2025stepguitechnicalreport}, and GUI-OWL-1.5-8B~\citep{ye2025mobileagentv3fundamentalagentsgui}, \textbf{(II) proprietary VLMs} coupled with a grounding module, following UGround~\citep{gou2025navigatingdigitalworldhumans} and GUI-Libra~\citep{yang2026guilibra}, including GPT-4o~\citep{openai2024gpt4o}, GPT-4.1, GPT-5-mini, and GPT-5~\citep{openaiGPT5Here}. We also incorporate reported or reproduced results from prior studies~\citep{liu2025scalecua,yang2026guilibra}, and \textbf{(III) memory-augmented GUI agents}, including Agent Workflow Memory (AWM)~\citep{wang2024agentworkflowmemory}, ReasoningBank~\citep{ouyang2026reasoningbankscalingagentselfevolving}, CoMEM~\citep{wu2025autoscalingcontinuousmemorygui} and HyMEM~\citep{zhu2026hybrid}.

\subsection{Main Results}
\vspace{-0.4em}

\begin{figure*}[t]
\centering

\begin{minipage}[!t]{0.47\textwidth}
    \centering
    \includegraphics[width=\linewidth]{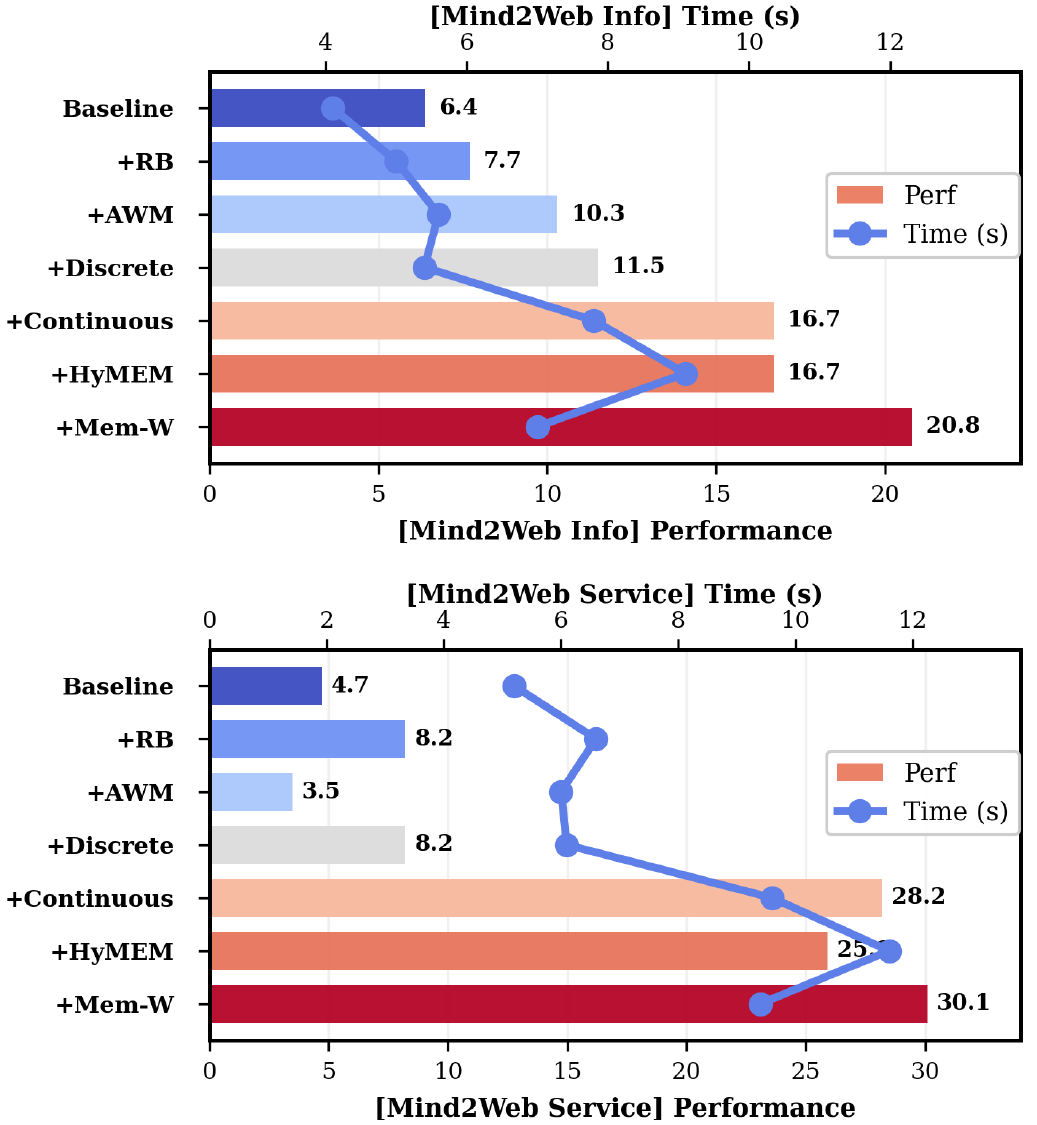}
    \caption{\small Comparison of memory-enhanced models. ``RB'' denotes ReasoningBank. }
    \label{fig:mem_compare}
\end{minipage}
\hfill
\begin{minipage}[!t]{0.47\textwidth}
    \centering
    \scriptsize
    \vspace{-2em}
    \captionof{table}{\small Performance comparison on MMInA and Mind2Web subsets. The metric used is success rate.}
    \vspace{-0.1em}
    \label{tab:mmina_mind2web_memory}
    \setlength{\tabcolsep}{3pt}
\begin{tabular}{llll}
\toprule
\multirow{2}{*}{\textbf{Model}}
& \textbf{MMInA}
& \multicolumn{2}{c}{\textbf{Mind2Web}} \\
\cmidrule(lr){2-2} \cmidrule(lr){3-4}
& Shop
& Info & Service \\
\midrule

\multicolumn{4}{c}{\textbf{Open-source Native Models}} \\
\midrule
Qwen2.5-VL-7B
& 16.00 & 9.80 & 17.65 \\

GLM-4.1V-9B-Thinking
& 28.50 & 10.78 & 26.47 \\

GUI-Owl-1.5-8B
& 5.00 & 8.82 & 30.39 \\

Qwen3-VL-2B
& 9.50 & 9.80 & 23.52 \\

Qwen3-VL-8B
& 19.50 & 13.73 & 26.47 \\

Qwen3-VL-32B
& 36.00 & 13.73 & 27.45 \\

UI-Venus-1.5-30B-A3B
& 13.50 & 3.92 & 8.82 \\

MAI-UI-2B
& 2.50 & 19.41 & 21.57 \\

Step-GUI-4B
& 19.50 & 12.75 & 21.56 \\

\midrule
\multicolumn{4}{c}{\textbf{\ourmethod Models}} \\
\midrule
UI-TARS-1.5-7B
& 5.50 & 5.88 & 6.86 \\

\rowcolor{LightCyan1}
\textbf{$\hookrightarrow$ \ourmethod-7B (Ours)}
& \textbf{32.50} {\tiny (+27.00)}
& \textbf{20.59} {\tiny (+14.71)} & \textbf{22.55} {\tiny (+15.69)} \\

UI-Venus-1.5-8B
& 18.50 & 5.88 & 15.69 \\

\rowcolor{LightCyan1}
\textbf{$\hookrightarrow$ \ourmethod-8B (Ours)}
& \textbf{48.50} {\tiny (+30.00)}
& \textbf{23.53} {\tiny (+17.65)}
& \textbf{36.27} {\tiny (+20.58)} \\

Qwen3-VL-4B
& 11.50 & 13.72 & 14.71 \\

\rowcolor{LightCyan1}
\textbf{$\hookrightarrow$ \ourmethod-4B (Ours)}
& \textbf{40.50} {\tiny (+29.00)}
& \textbf{22.55} {\tiny (+8.83)}
& \textbf{26.47} {\tiny (+11.76)} \\

\bottomrule
\end{tabular}
\end{minipage}
\vspace{-0.5em}
\end{figure*}

Tables~\ref{tab:ac_high_low_odyssey_merged} and~\ref{tab:mmina_mind2web_memory} show that \ourmethod consistently improves frozen GUI backbones across both \textit{mobile control} and \textit{web navigation} benchmarks.
On AndroidControl-v2, \ourmethod improves Qwen3-VL-4B by $+13.77$/$+11.57$ on AC-v2-High Pass@1/Pass@4 and by $+6.02$/$+6.80$ on AC-v2-Low, while improving UI-Venus-1.5-8B by $+7.53$/$+10.30$ and $+6.28$/$+7.03$ on the same settings.
On GUI-Odyssey, \ourmethod further improves Qwen3-VL-4B by $+2.18$/$+2.20$ on TM/EM and UI-Venus-1.5-8B by $+1.76$/$+2.60$.
The gains are especially strong on web navigation: on MMINA, \ourmethod raises UI-TARS-1.5-7B, UI-Venus-1.5-8B, and Qwen3-VL-4B by $+27.00$, $+30.00$, and $+29.00$ points, respectively; on Multimodal-Mind2Web, it improves UI-TARS-1.5-7B by $+14.71$/$+15.69$, UI-Venus-1.5-8B by $+17.65$/$+20.58$, and Qwen3-VL-4B by $+8.83$/$+11.76$ on Info/Service.
Notably, \ourmethod-8B reaches $94.22$ Pass@4 on AC-v2-Low, exceeding GPT-5+UGround and GUI-Libra-8B, while also lifting modest backbones to levels competitive with substantially larger open-source GUI models.
These results indicate that the bottleneck in long-horizon GUI agency is not merely model scale, but how interaction history and prior experience are represented and exposed to the policy.

\vspace{-0.4em}
\subsection{Comparison With Memory-augmented Agents}
\vspace{-0.4em}

To further compare \ourmethod with existing memory-augmented GUI agents, we evaluate several representative memory designs on Multimodal-Mind2Web in \Cref{fig:mem_compare}.
We include three discrete-token memory baselines, namely ReasoningBank~\citep{ouyang2026reasoningbankscalingagentselfevolving}, AWM~\citep{wang2024agentworkflowmemory}, and a graph-based discrete memory adapted from HyMEM~\citep{zhu2026hybridselfevolvingstructuredmemory}, which instantiate different explicit memory structures, such as reusable execution templates distilled from successful trajectories or graph-organized experience, but all expose memory to the GUI agent as discrete textual tokens.
We also compare with a continuous latent-memory baseline adapted from CoMEM~\citep{wu2025autoscalingcontinuousmemorygui}, as well as HyMEM~\citep{zhu2026hybridselfevolvingstructuredmemory}, which combines readable symbolic strategies with embedding-based trajectory memory.

\vspace{-0.4em}
\paragraph{Results.}
As shown in \Cref{fig:mem_compare}, \ourmethod achieves the strongest performance among all memory-augmented agents on both Multimodal-Mind2Web subsets.
On the Info subset, \ourmethod improves the baseline from $6.4$ to $20.8$, outperforming both CoMEM ($16.7$) and HyMEM ($16.7$).
On the Service subset, \ourmethod reaches $30.1$, improving over the baseline by $+25.4$ points and surpassing Continuous memory ($28.2$) and HyMEM ($25.9$).
Why can \ourmethod outperform CoMEM and HyMEM, even though they also use latent embeddings as GUI memory carriers?
We attribute this advantage to two factors: first, their working memory remains relatively coarse, largely relying on truncated nearest-neighbor screenshots, whereas \ourmethod dynamically distills historical trajectories into compact, decision-relevant latent summaries (\Cref{eq:working-latents}); second, their training signal is mainly expert-trajectory imitation, while \ourmethod introduces outcome-aware supervision (\Cref{eq:outcome-aware-rl}), enabling memory to preserve long-horizon procedural evidence that is actually tied to task success.
In terms of efficiency, \ourmethod also remains lightweight: it is faster than HyMEM and CoMEM on Mind2Web, while delivering higher performance.

\begin{figure*}[t]
\centering

\begin{minipage}[t]{0.60\textwidth}
\centering
\vspace{0pt}
\scriptsize
\captionof{table}{\small Ablation and efficiency analysis on MMINA.
Metrics include success rate (Succ.), average number of executed steps per task (\#Steps), the percentage of tasks reaching the maximum-step limit (Hit-Max), average runtime per task/step.}
\label{tab:mmina_ablation}
\vspace{-0.5em}
\setlength{\tabcolsep}{3pt}
\begin{tabular}{llccccc}
\toprule
\textbf{Backbone} & \textbf{Setting} 
& \textbf{Succ.} 
& \textbf{\#Steps} 
& \textbf{Hit-Max} 
& \textbf{Time/Task} 
& \textbf{Time/Step} \\
\midrule

\multirow{4}{*}{\venusmodel UI-Venus}
& Vanilla 
& 18.5 & 13.0 & 70.5 & 83.2 & 6.4 \\
& w/o Working 
& 47.5 & 7.7 & 28.5 & 82.4 & 10.7 \\
& w/o Experiential 
& 43.0 & 10.0 & 52.1 & 142.0 & 14.2 \\
& Full \ourmethod 
& \cellcolor{blue!10}\textbf{48.5} & 8.2 & 42.2 & 127.9 & 15.6 \\

\midrule

\multirow{4}{*}{\qwenmodel Qwen}
& Vanilla 
& 11.5 & 15.0 & 99.0 & 102.0 & 6.8 \\
& w/o Working 
& 38.5 & 8.3 & 33.0 & 105.4 & 12.7 \\
& w/o Experiential 
& 35.5 & 10.3 & 52.1 & 139.1 & 13.5 \\
& Full \ourmethod 
& \cellcolor{blue!5}40.5 & 7.1 & 22.0 & 117.9 & 16.6 \\

\midrule
\multicolumn{7}{l}{\color{gray}\textit{GUI Agent Baselines}} \\

\multicolumn{2}{l}{GLM-4.1V-9B-Thinking}
& 28.5 & 10.5 & 53.8 & 180.6 & 17.2 \\

\multicolumn{2}{l}{Qwen3-VL-32B}
& 36.0 & 12.7 & 59.0 & 125.7 & 9.9 \\

\multicolumn{2}{l}{Step-GUI-4B}
& 19.5 & 2.5 & 5.5 & 32.3 & 12.9 \\

\multicolumn{2}{l}{UI-TARS-1.5-7B}
& 5.5 & 13.3 & 75.5 & 81.1 & 6.1 \\

\multicolumn{2}{l}{UI-Venus-1.5-30B-A3B}
& 13.5 & 9.6 & 32.0 & 119.0 & 12.4 \\

\bottomrule
\end{tabular}
\end{minipage}
\hfill
\begin{minipage}[t]{0.37\textwidth}
\centering
\vspace{0pt}
\includegraphics[width=\linewidth]{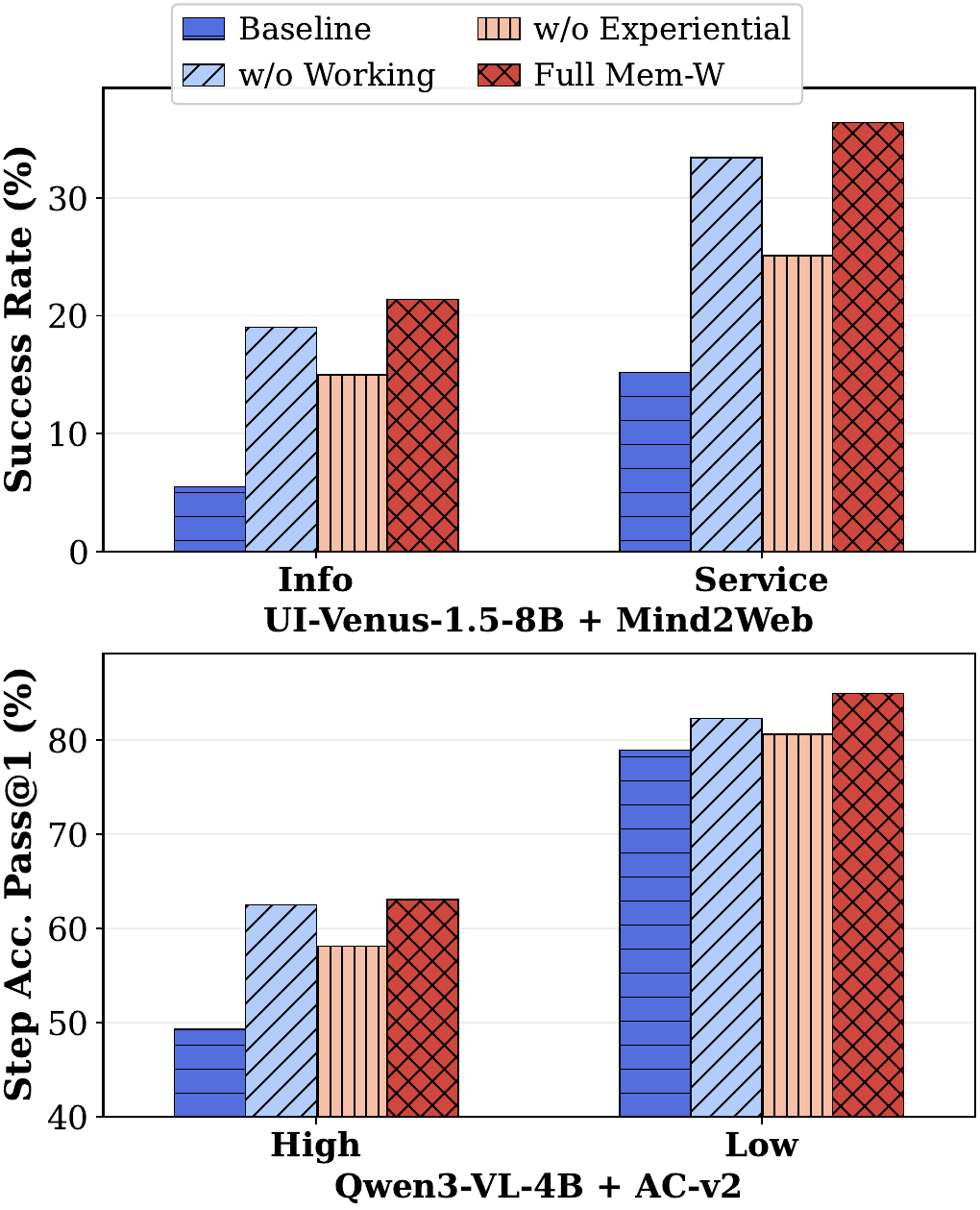}
\vspace{-1.6em}
\captionof{figure}{\small Inference-time ablation.}
\label{fig:inference_decomposition}
\end{minipage}
\vspace{-1.2em}
\end{figure*}

\vspace{-0.4em}
\subsection{Ablation Study}

\vspace{-0.4em}
We analyze the contribution of each memory source by selectively removing it from the latent context in \Cref{eq:weaved-policy}: \textit{w/o Working} sets $\mathbf{Z}^{\mathrm{work}}_t=\varnothing$, leaving only retrieved experiential memory and the local raw context, while \textit{w/o Experiential} sets $\mathbf{Z}^{\mathrm{proc}}_t=\varnothing$, leaving compressed in-episode history and the current GUI context. 
As shown in \Cref{fig:inference_decomposition} and \Cref{tab:mmina_ablation}, the full \ourmethod configuration consistently performs best across backbones and benchmarks: on MMINA, it improves UI-Venus from $18.5$ to $48.5$ and Qwen from $11.5$ to $40.5$, while reducing Hit-Max (the rate of hitting max step limit) from $70.5\%$ to $42.2\%$ and from $99\%$ to $22\%$, respectively. 
These results suggest that working memory and experiential memory are complementary rather than redundant: the former preserves unfinished in-session progress, while the latter supplies reusable procedural evidence from prior trajectories.

\vspace{-0.4em}
\subsection{Efficiency Analysis}
\vspace{-0.4em}
While constructing experiential and working memory requires an additional compression step, the overall runtime remains within the same practical range as representative GUI agents, and the added computation is accompanied by substantially more reliable task completion. As shown in \Cref{tab:mmina_ablation}, full \ourmethod achieves the best success rate among all compared settings, while reducing the tendency to exhaust the maximum-step budget, indicating that latent memory helps the agent reach successful trajectories more decisively rather than merely extending interaction length. Compared with stronger open-source baselines, \ourmethod also provides a favorable accuracy--efficiency trade-off: for example, it surpasses GLM-4.1V-9B-Thinking and Qwen3-VL-32B in success rate while maintaining comparable or lower task-level runtime. These results suggest that the memory compressor introduces a controlled and worthwhile computational cost.

\vspace{-0.4em}
\subsection{Framework Analysis}\label{sec:exp-analysis}
\vspace{-0.4em}
Beyond aggregate performance, we further analyze how \ourmethod behaves under two core design factors of the latent-memory framework.
We first study the impact of the number of retrieved trajectories, which controls how much experiential evidence is exposed to the compressor at inference time.
We then vary the size of the external memory bank $\mathcal{B}$ to examine whether a richer trajectory reservoir translates into stronger downstream GUI performance.

\begin{table*}[h]
\centering
\scriptsize

\begin{minipage}[!h]{0.52\textwidth}
\centering
\vspace{0pt}
\includegraphics[width=\linewidth]{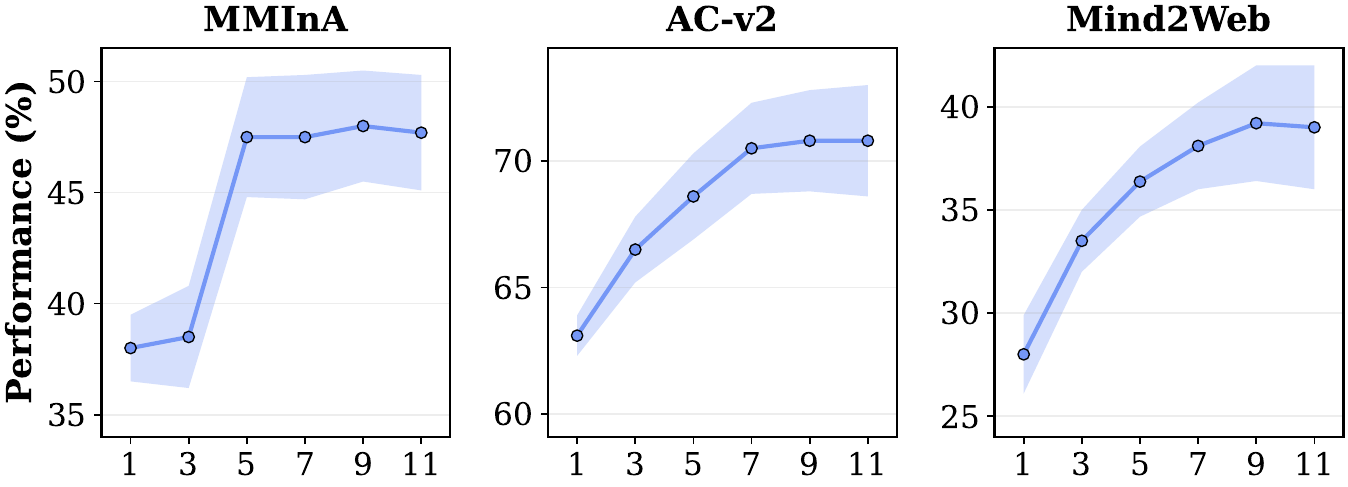}
\vspace{-1.5em}
\captionof{figure}{\small Impact of the number of retrieved trajectories. We vary the retrieval budget $M$ in \Cref{eq:procedural-retrieval} and report the corresponding performance. The model is UI-Venus.}
\label{fig:impact_retrieved_trajectories}
\end{minipage}
\hfill
\begin{minipage}[!h]{0.43\textwidth}
\centering
\vspace{-15pt}
\captionof{table}{\small Impact of memory bank size. We vary the number of trajectories stored in $\mathcal{B}$ and evaluate whether larger experiential reservoirs improve latent procedural memory.}
\label{tab:impact_memory_bank_size}
\vspace{-0.5em}
\setlength{\tabcolsep}{4pt}
\begin{tabular}{ll|cccc}
\toprule
\textbf{Model} & \textbf{Bench} & $10\mathrm{K}$ & $20\mathrm{K}$ & $30\mathrm{K}$ & $50\mathrm{K}$ \\
\midrule
\qwenmodel Qwen & MMINA & 35.00 & 38.00 & 39.50 & 40.00 \\
\qwenmodel Qwen & Mind2Web & 25.49 & 26.47 & 27.45 & 28.43 \\
\venusmodel UI-Venus & MMINA & 37.00 & 43.00 & 47.50 & 50.50 \\
\venusmodel UI-Venus & Mind2Web & 29.41 & 33.33 & 36.27 & 38.24 \\
\bottomrule
\end{tabular}
\end{minipage}
\vspace{-0.5em}
\end{table*}

\vspace{-0.5em}
\paragraph{Impact of Retrieved Trajectories.}\Cref{fig:impact_retrieved_trajectories} shows the effect of varying the retrieval budget $M$. Across MMINA, AC-v2, and Mind2Web, performance generally improves as more trajectories are retrieved, indicating that the procedural memory block $\mathbf{Z}^{\mathrm{proc}}_t$ benefits from richer experiential evidence.On MMINA, performance rises from $38.0$ at $M=1$ to $47.5$ at $M=5$, and remains stable thereafter.On AC-v2, the score increases from $63.1$ to $70.8$ as $M$ grows from $1$ to $9$, while Mind2Web improves from $28.0$ to $39.2$ over the same range.The gains gradually saturate for larger $M$, suggesting that \ourmethod can exploit additional trajectories, but excessive retrieval provides diminishing returns once the most relevant procedural evidence has been covered.

\vspace{-0.5em}
\paragraph{Impact of Memory Bank Size.}
\Cref{tab:impact_memory_bank_size} studies how the scale of the external trajectory bank $\mathcal{B}$ affects latent procedural memory.
Across all four settings, enlarging $\mathcal{B}$ from $10\mathrm{K}$ to $50\mathrm{K}$ brings clear and consistent gains: Qwen3-VL-4B improves by $+5.00$ on MMINA and $+2.94$ on Mind2Web, while UI-Venus improves by $+13.50$ on MMINA and $+8.83$ on Mind2Web.
The effect is especially pronounced for UI-Venus, where performance rises from $37.00$ to $50.50$ on MMINA and from $29.41$ to $38.24$ on Mind2Web, showing that \ourmethod can continue to benefit from a richer experiential reservoir at larger scales.
Even for Qwen3-VL-4B, the gains remain steady as the bank grows, with MMINA increasing from $35.00$ at $10\mathrm{K}$ to $39.50$ at $30\mathrm{K}$ and further to $40.00$ at $50\mathrm{K}$.
These results suggest that \ourmethod can effectively exploit larger trajectory banks, with the learned compressor transforming additional retrieved evidence into compact and informative latent memory.

\subsection{Case Study}
We provide qualitative case studies to illustrate how \ourmethod uses latent memory in long-horizon GUI interaction.
\Cref{fig:case1} and \Cref{fig:case2} show web and mobile examples, respectively, where the agent must preserve task constraints, intermediate progress, and executable action state across multiple steps.
\Cref{fig:case2-exp} further visualizes the retrieved trajectories for the mobile case, showing how experiential memory complements in-session working memory with reusable procedural evidence.

\vspace{-0.4em}
\section{Conclusion}
\vspace{-0.4em}
This work introduced \ourmethod, a framework for latent-memory-native GUI agents that unifies working memory and
experiential memory in a single continuous context space, replacing
prescribed memory taxonomies with a shared, end-to-end learnable
trajectory-to-latent compressor. By projecting both in-session history and
cross-session experience through the same mechanism, \ourmethod lets
task-relevant memory structure emerge from training rather than manual
design. Experiments across four web and mobile benchmarks show that this
unified latent interface consistently improves diverse backbones and
outperforms agents equipped with hand-engineered memory architectures.
More broadly, these results suggest that latent context can be a practical foundation component for future GUI agent architectures: a common substrate where perception, memory, procedural experience, and ongoing task state can be organized in the same machine-native form used by the policy itself.
We hope this perspective opens a path toward interactive multimodal agents whose long-horizon competence is built not around externally prescribed memory modules, but around learned latent context interfaces.

\section*{Contributions}

\begin{itemize}
    \item \textbf{Core Contributors}: Guibin Zhang, Yaohui Ling
    \item \textbf{Contributors}: Fanci Meng
    \item \textbf{Corresponding Authors}: Kun Wang, Shuicheng Yan
\end{itemize}

If you have any questions regarding the code, paper details, or other aspects of this work, you are very welcome to contact the authors at \url{guibinz@outlook.com} or via raising a \href{https://github.com/bingreeky/MemEvolve}{Github} issue. 

\clearpage

\bibliographystyle{plainnat}
\bibliography{refs}

\appendix

\section{Training Details}\label{app:train}

This appendix provides the full training procedure for \ourmethod, expanding on
the two-stage recipe described in \Cref{sec:method-training}. We restate the
key convention: the GUI agent $\bm{\Pi}_\theta$ (encoder
$\mathbf{G}_\theta$, decoder, and all associated weights) is \emph{frozen}
throughout both stages. Only the compressor parameters
$\phi=\{\mathbf{Q},\operatorname{QFormer}_\phi,\mathbf{P}_\phi,
\gamma,\mathbf{b}^{\mathrm{proc}},\mathbf{b}^{\mathrm{work}}\}$ receive
gradient updates.

\subsection{Stage 1: Self-Distillation}\label{app:stage1}

\paragraph{Data construction.}
We sample a trajectory $\tau=(x,s_{1:T},y)$ from a demonstration dataset
$\mathcal{D}$ and uniformly draw a timestep $t$ such that the expired prefix
$\bar{s}_t=s_{1:\max(0,\,t-L-1)}$ is non-empty (i.e.\ $t>L+1$), so that
working-memory compression is exercised. The raw context window is
$\rho_t=s_{\max(1,t-L):t-1}$. To also exercise the procedural-memory path, we
retrieve $M$ trajectories from the external bank $\mathcal{B}$ using
\Cref{eq:procedural-retrieval}, yielding the retrieved index set
$\mathcal{I}_t$.

\paragraph{Teacher context.}
The teacher branch feeds the same frozen agent an extended raw window of $L'$
steps ($L'\gg L$):
\begin{equation}
  \chi_t^{\mathrm{raw}}
  =
  \bigl(x,\;s_{\max(1,\,t-L'):t-1},\;o_t\bigr).
  \label{eq:app-teacher-context}
\end{equation}
Because $\bm{\Pi}_\theta$ is frozen and $\chi_t^{\mathrm{raw}}$ contains no
learnable latent tokens, the teacher output
$\bm{\Pi}_\theta(\cdot\mid\chi_t^{\mathrm{raw}})$ is treated as a fixed target
(gradients are stopped via $\operatorname{sg}[\cdot]$). Note that the teacher
observes a larger but still finite window; it is not assumed to see the entire
trajectory.

\paragraph{Student context.}
The student branch constructs the weaved input $\mathbf{U}_t$ from
\Cref{eq:weaved-policy}, with both procedural and working memory active.
The procedural-memory block is built by compressing the retrieved trajectories
with their outcome labels,
\begin{equation}
  \mathbf{Z}^{\mathrm{proc}}_t
  =
  \left[
  C_\phi(s^i_{1:T_i},y_i)
  \right]_{i\in\mathcal{I}_t},
  \label{eq:app-proc-context}
\end{equation}
while the working-memory block $\mathbf{Z}^{\mathrm{work}}_t$ is built by
partitioning the expired prefix $\bar{s}_t$ into non-overlapping chunks of
$W$ steps and compressing each chunk with the unknown-outcome marker
$\varnothing$ as in \Cref{eq:working-latents}. The student input is therefore
\begin{equation}
  \mathbf{U}_t^{\text{s1}}
  =
  \bigl[\,
    \mathbf{Z}^{\mathrm{proc}}_t
    + \mathbf{1}_{MK}(\mathbf{b}^{\mathrm{proc}})^\top
    \;;\;
    \mathbf{Z}^{\mathrm{work}}_t
    + \mathbf{1}_{J_tK}(\mathbf{b}^{\mathrm{work}})^\top
    \;;\;
    \mathbf{E}_\theta(x,o_t,\rho_t)
  \,\bigr].
  \label{eq:app-student-context}
\end{equation}

\paragraph{Loss.}
For each sample $(\tau,t)$ the stage-one loss is
\begin{equation}
  \mathcal{L}_{\mathrm{sd}}(\phi;\tau,t)
  =
  \underbrace{
    \ell_{\mathrm{gui}}\!\bigl(
      \bm{\Pi}_\theta(\cdot\mid\mathbf{U}_t^{\text{s1}}),\;a_t^\star
    \bigr)
  }_{\text{action loss}}
  \;+\;
  \lambda\;
  \underbrace{
    D_{\mathrm{KL}}\!\Bigl(
      \operatorname{sg}\bigl[
        \bm{\Pi}_\theta(\cdot\mid\chi_t^{\mathrm{raw}})
      \bigr]
      \;\Big\Vert\;
      \bm{\Pi}_\theta(\cdot\mid\mathbf{U}_t^{\text{s1}})
    \Bigr)
  }_{\text{distillation loss}},
  \label{eq:app-sd-loss}
\end{equation}
where $a_t^\star$ is the ground-truth action at step $t$ and $\lambda$ is a
balancing coefficient.

The action loss $\ell_{\mathrm{gui}}$ is the standard next-token
cross-entropy over the action token sequence: if the target action serialization
is $(a_t^{(1)},\ldots,a_t^{(S)})$, then
\begin{equation}
  \ell_{\mathrm{gui}}\!\bigl(
    \bm{\Pi}_\theta(\cdot\mid\mathbf{U}_t^{\text{s1}}),\;a_t^\star
  \bigr)
  =
  -\frac{1}{S}\sum_{s=1}^{S}
  \log p_{\theta,\phi}\!\bigl(a_t^{(s)}\mid\mathbf{U}_t^{\text{s1}},a_t^{(1:s-1)}\bigr),
  \label{eq:app-action-ce}
\end{equation}
where $p_{\theta,\phi}$ denotes the next-token probability of the frozen agent
conditioned on the latent-augmented context. Since $\theta$ is frozen, gradients
from both losses reach learnable parameters only through the procedural and
working latent blocks produced by $C_\phi$.

\subsection{Stage 2: RLOO-Based Outcome-Aware Optimization}\label{app:stage2}

Stage~2 keeps both procedural and working memory active and further optimizes
the compressor with environment feedback. The agent backbone $\theta$ remains
frozen; gradients reach $\phi$ through the differentiable latent tokens in the
weaved context $\mathbf{U}_t$.

\paragraph{Rollout and reward.}
For each task instruction $x$ drawn from a task distribution $p(x)$, we sample
$G$ independent rollout trajectories from the current policy:
\begin{equation}
  \tau^{(g)}
  =
  \bigl(x,\,s^{(g)}_{1:T^{(g)}},\,y^{(g)}\bigr),
  \qquad
  g=1,\ldots,G,
  \qquad
  \tau^{(g)}\sim\pi_{\theta,\phi},
  \label{eq:app-rollouts}
\end{equation}
where each trajectory is produced by conditioning the frozen agent on the
latent-augmented context $\mathbf{U}_t^{(g)}$ constructed with the current
compressor $C_\phi$. The terminal reward is binary:
$R(\tau^{(g)})\in\{0,1\}$, with $1$ indicating task success.

\paragraph{RLOO advantage estimation.}
Following RLOO
estimator~\citep{ahmadian2024basicsrevisitingreinforcestyle}, we construct an unbiased,
low-variance advantage for each rollout by using the other $G-1$ rollouts to
the same instruction as a baseline. For the $g$-th rollout, the advantage is
\begin{equation}
  \hat{A}^{(g)}
  =
  R\!\bigl(\tau^{(g)}\bigr)
  \;-\;
  \frac{1}{G-1}
  \sum_{\substack{g'=1\\g'\neq g}}^{G}
  R\!\bigl(\tau^{(g')}\bigr).
  \label{eq:app-rloo-advantage}
\end{equation}
This per-instruction baseline is parameter-free and does not require a learned
value function. For computational efficiency, we rewrite it as
\begin{equation}
  \hat{A}^{(g)}
  =
  \frac{G}{G-1}
  \left(
    R\!\bigl(\tau^{(g)}\bigr)
    -
    \bar{R}_x
  \right),
  \qquad
  \bar{R}_x
  =
  \frac{1}{G}\sum_{g'=1}^{G}R\!\bigl(\tau^{(g')}\bigr),
  \label{eq:app-rloo-efficient}
\end{equation}
where $\bar{R}_x$ is the mean reward over all $G$ rollouts for instruction $x$
and is computed once per instruction.

\paragraph{Per-trajectory policy gradient.}
Because the reward is trajectory-level (success or failure of the entire
episode), we assign the same advantage $\hat{A}^{(g)}$ to every timestep within
rollout $\tau^{(g)}$. The per-trajectory policy-gradient loss for the $g$-th
rollout is
\begin{equation}
  \mathcal{L}_{\mathrm{pg}}^{(g)}(\phi)
  =
  -\hat{A}^{(g)}
  \;\cdot\;
  \frac{1}{T^{(g)}}
  \sum_{t=1}^{T^{(g)}}
  \log\bm{\Pi}_\theta\!\bigl(a_t^{(g)}\mid\mathbf{U}_t^{(g)}\bigr),
  \label{eq:app-pg-loss}
\end{equation}
where the log-probability is summed over the action tokens at each step and
averaged over timesteps.

\paragraph{KL regularization.}
To prevent the compressor from drifting too far from the stage-one solution, we
add a KL penalty against a frozen reference policy $\pi_{\mathrm{ref}}$. The
reference policy uses the stage-one compressor $C_{\phi_{\mathrm{ref}}}$ (with
parameters frozen after stage~1) to construct a reference context
$\mathbf{U}^{\mathrm{ref}}_t$, while the agent decoder $\bm{\Pi}_\theta$
remains the same. The per-step KL term is
\begin{equation}
  D_t^{(g)}
  =
  D_{\mathrm{KL}}\!\Bigl(
    \bm{\Pi}_\theta\!\bigl(\cdot\mid\mathbf{U}_t^{(g)}\bigr)
    \;\Big\Vert\;
    \bm{\Pi}_\theta\!\bigl(\cdot\mid\mathbf{U}_t^{\mathrm{ref},(g)}\bigr)
  \Bigr),
  \label{eq:app-kl-step}
\end{equation}
where $\mathbf{U}_t^{\mathrm{ref},(g)}$ is constructed identically to
$\mathbf{U}_t^{(g)}$ except that the latent tokens are produced by the frozen
$C_{\phi_{\mathrm{ref}}}$ instead of the current $C_\phi$. Because
$\mathbf{U}_t^{\mathrm{ref},(g)}$ involves no learnable parameters, gradients
of $D_t^{(g)}$ flow only into $\phi$ through $\mathbf{U}_t^{(g)}$.

\paragraph{Combined objective.}
The full stage-two objective, averaged over a batch of instructions and their
rollouts, is
\begin{equation}
  \mathcal{L}_{\mathrm{s2}}(\phi)
  =
  \mathbb{E}_{x\sim p(x)}
  \left[
    \frac{1}{G}\sum_{g=1}^{G}
    \left(
      \mathcal{L}_{\mathrm{pg}}^{(g)}(\phi)
      \;+\;
      \frac{\beta}{T^{(g)}}
      \sum_{t=1}^{T^{(g)}}
      D_t^{(g)}
    \right)
  \right],
  \label{eq:app-stage2-full}
\end{equation}
where $\beta$ is the KL penalty coefficient. Expanding the policy-gradient and
KL terms together, the per-rollout contribution can equivalently be written as
\begin{equation}
  \frac{1}{T^{(g)}}
  \sum_{t=1}^{T^{(g)}}
  \left(
    -\hat{A}^{(g)}\;
    \log\bm{\Pi}_\theta\!\bigl(a_t^{(g)}\mid\mathbf{U}_t^{(g)}\bigr)
    \;+\;
    \beta\;
    D_{\mathrm{KL}}\!\Bigl(
      \bm{\Pi}_\theta(\cdot\mid\mathbf{U}_t^{(g)})
      \;\Big\Vert\;
      \bm{\Pi}_\theta(\cdot\mid\mathbf{U}_t^{\mathrm{ref},(g)})
    \Bigr)
  \right),
  \label{eq:app-stage2-expanded}
\end{equation}
which directly corresponds to \Cref{eq:outcome-aware-rl} in the main text, with
$\hat{R}(\tau)$ instantiated by the RLOO advantage $\hat{A}^{(g)}$.

At each step $t$, the frozen decoder $\bm{\Pi}_\theta$ receives
$\mathbf{U}_t^{(g)}$ as input embeddings. The latent blocks
$\mathbf{Z}^{\mathrm{proc}}_t$ and $\mathbf{Z}^{\mathrm{work}}_t$ inside
$\mathbf{U}_t^{(g)}$ are differentiable outputs of the compressor $C_\phi$.
When computing $\nabla_\phi\mathcal{L}_{\mathrm{s2}}$, the gradient passes from
the scalar loss through the frozen decoder's attention layers (which are applied
but not updated) into the latent tokens, and then into the QFormer and
projection head of $C_\phi$. The retrieval operation
($\operatorname{Top\text{-}M}$ over frozen encoder outputs) is discrete and
contributes no gradient; only the compression of the retrieved segments is
differentiable. Similarly, the reference context
$\mathbf{U}_t^{\mathrm{ref},(g)}$ is fully detached, so the KL term's gradient
flows only through the current compressor's output.

\section{Experiment Details}
\subsection{Training Dataset}\label{app:training-data}

\paragraph{Web Training Dataset.}
The Web training dataset is mainly from CoMEM~\citep{wu2025autoscalingcontinuousmemorygui}; it covers 13 domains, including academic, education, finance, government, health, shopping, tech, and travel. Each sample is labeled as either successful or failed according to its task execution outcome. In total, the Web training set contains 81,527 samples, consisting of 11,176 successful trajectories and 70,351 failed trajectories. During training, we use the successful trajectories as training examples.

\begin{table}[t]
\centering
\caption{Web Training Data}
\label{tab:web_training_data}
\small
\renewcommand{\arraystretch}{1.15}
\setlength{\tabcolsep}{8pt}
\begin{tabular}{lrrrr}
\toprule
\textbf{Domain} & \textbf{Success} & \textbf{Failure} & \textbf{Total} & \textbf{Success Rate (\%)} \\
\midrule
academic      & 1246 &  6579 &  7825 & 15.92 \\
education     & 1045 &  6536 &  7581 & 13.78 \\
entertainment &  297 &  1817 &  2114 & 14.05 \\
finance       &  692 &  3907 &  4599 & 15.05 \\
food          &  835 &  4840 &  5675 & 14.71 \\
government    & 1417 & 10491 & 11908 & 11.90 \\
health        & 1380 &  9447 & 10827 & 12.75 \\
news          &  600 &  3386 &  3986 & 15.05 \\
service       &  861 &  5530 &  6391 & 13.47 \\
shopping      &  612 &  4634 &  5246 & 11.67 \\
social        &  415 &  1666 &  2081 & 19.94 \\
tech          & 1505 &  7231 &  8736 & 17.23 \\
travel        &  271 &  4287 &  4558 &  5.95 \\
\midrule
\textbf{Overall Summary} & \textbf{11176} & \textbf{70351} & \textbf{81527} & \textbf{13.71} \\
\bottomrule
\end{tabular}
\end{table}

\paragraph{Web Memory Bank.}

The web memory bank is also sampled from \citep{wu2025autoscalingcontinuousmemorygui}; it is designed to store reusable task-execution trajectories and domain-specific knowledge. It contains a total of 22,346 successful memories, covering the same 13 domains as the Web training set. These memories provide useful references for planning, retrieval, and decision-making when the model encounters similar Web-based tasks.

\begin{table}[t]
\centering
\caption{Web Memory Bank}
\label{tab:web_memory_bank}
\small
\renewcommand{\arraystretch}{1.15}
\setlength{\tabcolsep}{8pt}
\begin{tabular}{lrrrr}
\toprule
\textbf{Domain} & \textbf{Success} & \textbf{Failure} & \textbf{Total} & \textbf{Success Rate (\%)} \\
\midrule
academic      & 2491 & 13158 & 15649 & 15.92 \\
education     & 2089 & 13072 & 15161 & 13.78 \\
entertainment &  593 &  3634 &  4227 & 14.03 \\
finance       & 1384 &  7814 &  9198 & 15.05 \\
food          & 1670 &  9680 & 11350 & 14.71 \\
government    & 2834 & 20982 & 23816 & 11.90 \\
health        & 2760 & 18894 & 21654 & 12.75 \\
news          & 1199 &  6772 &  7971 & 15.04 \\
service       & 1722 & 11060 & 12782 & 13.47 \\
shopping      & 1223 &  9268 & 10491 & 11.66 \\
social        &  830 &  3332 &  4162 & 19.94 \\
tech          & 3010 & 14462 & 17472 & 17.23 \\
travel        &  541 &  8574 &  9115 &  5.94 \\
\midrule
\textbf{Overall Summary} & \textbf{22346} & \textbf{140702} & \textbf{163048} & \textbf{13.71} \\
\bottomrule
\end{tabular}
\end{table}

\paragraph{Mobile Training Dataset.}

The Mobile training dataset is drawn from GUI-Odyssey~\citep{guiodyssey}, using the official training split provided at \url{https://huggingface.co/datasets/OpenGVLab/GUI-Odyssey/blob/main/splits/random_split.json}.
We emphasize that training is conducted only on the official training set, while evaluation uses the official test set, ensuring that \textbf{there is no overlap or data leakage between training and evaluation}. It covers six categories of mobile tasks, including Utility Operations, Social Posting, Media Services, Data Organization, Cross-App Workflows, and Online Purchasing. The Mobile training set contains 2,610 samples, including 2,486 successful trajectories and 124 failed trajectories. Similar to the Web setting, we use the successful trajectories for training.

\begin{table}[t]
\centering
\caption{Mobile Training Data}
\label{tab:mobile_training_data}
\small
\renewcommand{\arraystretch}{1.15}
\setlength{\tabcolsep}{8pt}
\begin{tabular}{lrrrr}
\toprule
\textbf{Domain} & \textbf{Success} & \textbf{Failure} & \textbf{Total} & \textbf{Success Rate (\%)} \\
\midrule
utility               & 832 &  2 & 834 & 99.76 \\
social communication  & 532 &  0 & 532 & 100.00 \\
media services        & 517 &  3 & 520 & 99.42 \\
information handling  & 335 & 52 & 387 & 86.56 \\
cross-app workflow    & 232 & 19 & 251 & 92.43 \\
e-commerce            &  41 & 47 &  88 & 46.59 \\
\midrule
\textbf{Overall Summary} & \textbf{2489} & \textbf{123} & \textbf{2612} & \textbf{95.29} \\
\bottomrule
\end{tabular}
\end{table}

\paragraph{Mobile Memory Bank.}

The mobile memory bank is also constructed from the official GUI-Odyssey training split~\citep{guiodyssey}. It contains $4{,}972$ successful memories spanning all six categories of mobile tasks, capturing reusable operation patterns for cross-application workflows, information organization, and general mobile interaction. \textbf{We emphasize that the memory bank is built exclusively from training-set trajectories and has no overlap with the official test set used for evaluation, thereby avoiding any data leakage.}

\begin{table}[t]
\centering
\caption{Mobile Memory Bank}
\label{tab:mobile_memory_bank}
\small
\renewcommand{\arraystretch}{1.15}
\setlength{\tabcolsep}{8pt}
\begin{tabular}{lrrrr}
\toprule
\textbf{Domain} & \textbf{Success} & \textbf{Failure} & \textbf{Total} & \textbf{Success Rate (\%)} \\
\midrule
utility               & 1663 &   5 & 1668 &  99.70 \\
social communication  & 1063 &   0 & 1063 & 100.00 \\
media services        & 1033 &   6 & 1039 &  99.42 \\
information handling  &  669 & 104 &  773 &  86.55 \\
cross-app workflow    &  463 &  39 &  502 &  92.23 \\
e-commerce            &   81 &  95 &  176 &  46.02 \\
\midrule
\textbf{Overall Summary} & \textbf{4972} & \textbf{249} & \textbf{5221} & \textbf{95.23} \\
\bottomrule
\end{tabular}
\end{table}

\subsection{Parameter Configuration}\label{app:param}

During training, we adopt a parameter-efficient fine-tuning strategy based on LoRA, together with DeepSpeed ZeRO-3 for memory optimization and distributed training. The original parameters of both the language model backbone and the visual encoder are frozen, while only the LoRA adapters and the compressor-related modules are updated. This design reduces the overall training cost while preserving the general vision-language capabilities of the pretrained foundation model.

The model is trained using bfloat16 mixed precision, gradient checkpointing, a cosine learning-rate scheduler, and the AdamW optimizer. The learning rate is set to $5 \times 10^{-5}$, with a weight decay of $0.1$ and a warmup ratio of $0.03$. For LoRA, the rank is set to $16$, the scaling factor $\alpha$ is set to $32$, and the dropout rate is set to $0.05$. Under the default configuration, the per-GPU batch size is set to $2$. All the experiments are conducted on a server with 8 NVIDIA A800 (80GB) GPUs.

\subsection{Evaluation Setup}\label{app:evaluation-setup}

We evaluated our method on four benchmarks, namely MMInA, Mind2Web, GUI-Odyssey, and Android-Control-v2. As MMInA and Mind2Web share the same interactive GUI environment, we describe their experimental setup and results together. In contrast, the evaluations on GUI-Odyssey and Android-Control-v2 are presented separately.

\paragraph{Interactive GUI Evaluation on MMInA and Mind2Web.}
During the web evaluation stage, we deploy the agent in an interactive GUI environment and adopt a ReAct-style reasoning--acting paradigm. At each step, the agent receives the current web state as input, including the page screenshot, textual information extracted from the page, the task instruction, and the most recent action history. Based on this information, the model is required to analyze the current state, plan the next operation, and output a structured function call. The action space includes clicking, typing text, pressing keys, scrolling, waiting, and producing the final answer followed by termination.

For each task, we set the maximum number of interaction steps to 15. A task terminates when the agent explicitly outputs a \texttt{stop} action or when the maximum step limit is reached.

When continuous memory is enabled, the model is loaded and used for generation through a local Transformers backend. In this mode, the inference parameters are set to \texttt{temperature = 0.1}, \texttt{top\_p = 0.001}, and a maximum generation length of 10,000 new tokens, with KV-cache enabled. If no relevant memory is retrieved, the system falls back to the standard vLLM inference pipeline.

The memory module retrieves historical trajectories that are similar to the current task using a FAISS index. Specifically, embeddings are first normalized with L2 normalization, and retrieval is performed using inner-product similarity. In the actual evaluation, the top five most similar historical experiences are retrieved for each task and inserted into the system prompt as experience memory, providing demonstrations and contextual references for solving the current task.

\paragraph{Benchmark Evaluation on GUI-Odyssey.}

During each reasoning step in GUI-Odyssey, the model input consists of the current GUI observation, the task instruction, and the retrieved experience memory.

Since GUI-Odyssey is evaluated in an offline setting, the retrieval process differs from that used in interactive evaluation, where trajectory-level experiences can be directly retrieved during interaction. Instead, during inference, candidate experiences are retrieved from a step-level memory index. For each task, we retrieve the top five most similar historical experiences and incorporate them into the prompt as contextual references.

For generation, we adopt a deterministic decoding strategy. Specifically, the temperature is set to \texttt{0.0}, top-\textit{p} is set to \texttt{1.0}, and the maximum generation length for each single-step action prediction is set to \texttt{64} tokens.

\paragraph{Benchmark Evaluation on Android-Control-v2.}

For AndroidControl-v2, we evaluate models under two instruction settings: High-Level and Low-Level. In the High-Level setting, the model is provided only with the high-level task goal. It must infer the appropriate next action, such as where to click, what to type, whether to navigate back, or whether to scroll, based on the current screenshot and the interaction history. In the Low-Level setting, in addition to the high-level task goal, the model is given the oracle low-level instruction for the current step.

We further evaluate model performance under different numbers of generation attempts. Specifically, we consider two settings: Pass@1 and Pass@4. For Pass@1, a single output is generated for each sample, and the sample is considered successful only if this output is correct. For Pass@4, four candidate outputs are generated for each sample, and the sample is considered successful if any one of the four candidates is correct. We set the temperature to $0.0$ for Pass@1 and to $1.0$ for Pass@4, with a maximum generation length of $2048$ tokens.

Similar to GUI-Odyssey, during inference we retrieve candidate experiences from the step-level memory index. For each task, we retrieve the top-$5$ most similar historical experiences; the sensitivity analysis of this parameter has been provided in \Cref{sec:exp-analysis}.

\subsection{Evaluation Benchmark}

\paragraph{MMIna.}

MMInA is a multimodal, multi-hop, and open-ended benchmark for evaluating web agents in realistic web environments. It emphasizes the ability of an agent to autonomously plan action trajectories in a browser according to user instructions, while jointly leveraging webpage screenshots, HTML content, and multimodal information such as text and images for understanding and decision making.

The \textit{shopping} subset represents a typical e-commerce scenario in MMInA. The tasks all centered around products from OneStopMarket. The subset adopts a discrete web interaction action space, which contains 12 types of high-level actions. These actions cover the major operations required for completing shopping-related tasks in a web environment, including clicking, typing, and hovering over webpage elements; scrolling the page and issuing keyboard commands; as well as browser-level navigation actions such as visiting a URL, moving backward or forward, opening a new tab, closing a tab, and switching page focus, as detailed in \Cref{tab:mind2web_action_space}. In addition, the action space includes the \texttt{stop [answer]} action, which indicates task completion and returns the final answer.

\paragraph{Agent Inference Prompt.}
The prompt used during interactive GUI evaluation is dynamically constructed at each step. The system message is loaded from the agent prompt template and filled with the retrieved experience memory and the available tool specifications. The user-side context contains the current webpage screenshot, a generated page description, and the current task instruction. The simplified prompt template used during evaluation is shown below.

\begin{tcolorbox}[
    enhanced,
    breakable,
    colback=PromptBg,
    colframe=PromptOrange,
    boxrule=0.8pt,
    arc=2mm,
    title=\textbf{Agent Inference Prompt},
    colbacktitle=PromptOrange,
    coltitle=white,
    fonttitle=\bfseries,
    left=2mm,
    right=2mm,
    top=1mm,
    bottom=1mm
]

\begin{center}
\textbf{SYSTEM INSTRUCTION}
\end{center}

\vspace{1mm}

\begin{PromptVerbatim}
You are a GUI automation agent that can interact with web pages and applications using the ReAct (Reasoning and Acting) paradigm.

IMPORTANT: You MUST output your actions in structured JSON format that can be parsed directly. Use the function calling mechanism to execute actions.

Your task is to:
1. Analyze the current state of the page, including numerical labels on web elements.
2. Think through what needs to be done.
3. Determine the appropriate action to take.
4. Output the action in structured JSON format that can be parsed directly.

WORKFLOW GUIDELINES:
- If your previous action is type, then you must click related pages or scroll pages to find the information you need.
- When you need to search for information, directly type your search query, then click the search button.
- After clicking an element, if you need to interact with it further, such as typing, do so immediately.
- Do not repeat the same action multiple times. If something does not work, try a different approach.
- Always use function calling to execute actions. Do not describe actions in plain text.
- If the current page has no results, adjust your search term and try again.
- Pay attention to images, since many questions about shape, color, location, and visual content can be answered from screenshots.

ACTION GUIDELINES:
1. To input text, no need to click the textbox first. Directly type the content. After typing, the system automatically hits the ENTER key.
2. Distinguish between textboxes and search buttons. Do not type content into a button.
3. Execute only one action per iteration.
4. Strictly avoid repeating the same action if the webpage remains unchanged.
5. For complex tasks involving multiple questions or steps, select "stop" only at the very end.
6. Make sure all task requirements are satisfied before stopping.

WEB BROWSING GUIDELINES:
1. Do not interact with useless web elements such as Login, Sign-in, or donation buttons.
2. Visiting video websites is allowed, but the agent should not play videos.
3. Pay attention to filter and sorting functions, which can help solve conditions such as highest, cheapest, lowest, or earliest.
4. Pay attention to images and visual elements on the page.

EXAMPLE WORKFLOW:
{experience_memory}

Available actions:
{tools_section}

CRITICAL REQUIREMENTS:
1. Always use function calling. Never describe actions in plain text.
2. Provide clear reasoning in the reasoning parameter of each function call.
3. Be specific in descriptions to identify the correct elements.
4. Use proper JSON format for all function arguments.
5. Only execute one action at a time.
6. If an action fails, try a different approach or element.
7. For click and type actions, set a valid element_id corresponding to the numerical label of the target item.
8. For click and type actions, set valid coordinates in the format "<point>x1 y1</point>".
9. Use simple search terms when searching for information.
10. If the current page has no results, adjust the search term and try again.

CURRENT WEBPAGE OBSERVATION:
[Current webpage screenshot is provided as an image input.]

[Generated page description:]
{page_description}

FINAL PER-STEP TASK REMINDER:
**Current task:** {current_task}

IMPORTANT REMINDERS:
- Please specify the number label of the item you want to interact with, in the description of the action.
- Do not repeat the same action multiple times. Try different approaches if something does not work.
- If your previous action is type, then you must click related pages or scroll pages to find the information you need.
- Pay attention to images, since many questions about shape, color, location, and visual content can be answered from screenshots.
- If the current search term yields no results, adjust the search term and try again.
- You must provide an answer within the remaining steps.
- Only output one action at a time. Do not output multiple actions at once.

What would you like to do next?

Remember: Your responses must be structured function calls that can be parsed directly by the system.
\end{PromptVerbatim}

\end{tcolorbox}

\paragraph{Mind2Web.}

Mind2Web is designed to evaluate whether an agent can follow natural language instructions to complete cross-page, multi-step web operation tasks on real-world websites. The original Mind2Web benchmark contains 2,350 open-ended tasks, covering 137 websites across 31 domains. In our GUI-agent project, we adapt the original test set through a unified format conversion and an execution-interface wrapper. Specifically, the version of Mind2Web used in our experiments mainly provides task-level information, including the task identifier, start URL, user intent, website category, and evaluation configuration, but does not include complete human action trajectories.

For the action space, we do not directly rely on the trajectory-level actions from the original dataset. Instead, we integrate Mind2Web tasks into our interactive GUI environment. During evaluation, the agent is allowed to use the same action space as that used for MMInA, as described above. This unified action-space design enables Mind2Web and MMInA to be evaluated under the same GUI-agent execution framework, ensuring consistency in action representation, browser interaction, and task completion evaluation.

\begin{table}[H]
\centering
\caption{Unified Action Space Used for Mind2Web and MMInA Evaluation}
\label{tab:mind2web_action_space}
\small
\renewcommand{\arraystretch}{1.15}
\setlength{\tabcolsep}{8pt}
\begin{tabular}{lll}
\toprule
\textbf{Action} & \textbf{Action Format} & \textbf{Functionality} \\
\midrule
Click      & \texttt{click [element\_id]} 
           & Click a specified webpage element. \\
Type       & \texttt{type [element\_id] [text]} 
           & Enter text into a specified input field. \\
Hover      & \texttt{hover [element\_id]} 
           & Move the cursor over a webpage element. \\
Scroll     & \texttt{scroll [up/down]} 
           & Scroll the webpage upward or downward. \\
Press      & \texttt{press [key\_comb]} 
           & Execute a keyboard shortcut or key command. \\
Go to URL  & \texttt{goto [url]} 
           & Navigate directly to a specified URL. \\
Go Back    & \texttt{go\_back} 
           & Return to the previous page in browser history. \\
Go Forward & \texttt{go\_forward} 
           & Move forward in browser history. \\
New Tab    & \texttt{new\_tab} 
           & Open a new browser tab. \\
Close Tab  & \texttt{close\_tab} 
           & Close the current browser tab. \\
Page Focus & \texttt{page\_focus [page\_number]} 
           & Switch focus to a specified browser page or tab. \\
Stop       & \texttt{stop [answer]} 
           & Terminate the task and return the final answer. \\
\bottomrule
\end{tabular}
\end{table}

\paragraph{GUI-Odyssey.}

GUI-Odyssey is a dataset designed for mobile graphical user interface (GUI) agent tasks. It is primarily used to evaluate and train models on multi-step interaction capabilities in realistic mobile application environments. Each task in the dataset consists of a natural language instruction, a sequence of interface screenshots, and the corresponding action trajectory. Given the current screen state, an agent is expected to iteratively perform operations such as tapping, scrolling, text input, or system navigation in order to accomplish the user's goal. In the version used in our work, the action space mainly covers nine common types of mobile GUI interactions: \textit{Click}, \textit{Long Press}, \textit{Scroll}, \textit{Type}, \textit{Complete}, \textit{Impossible}, \textit{Home}, \textit{Back}, and \textit{Recent}, as detailed in \Cref{tab:gui_odyssey_action_space}.

\begin{table}[t]
\centering
\caption{Action Space of the GUI-Odyssey Dataset}
\label{tab:gui_odyssey_action_space}
\small
\renewcommand{\arraystretch}{1.15}
\setlength{\tabcolsep}{6pt}
\begin{tabular}{lll}
\toprule
\textbf{Action} & \textbf{Action Format} & \textbf{Functionality} \\
\midrule
Click      & \texttt{click [x,y]}              & Tap a screen position \\
Long Press & \texttt{long\_press [x,y]}        & Long-press a screen position \\
Scroll     & \texttt{scroll [x1,y1,x2,y2]}     & Swipe from one position to another \\
Type       & \texttt{type [text]}              & Enter text \\
Complete   & \texttt{complete}                 & Mark the task as completed \\
Impossible & \texttt{impossible}               & Mark the task as impossible \\
Home       & \texttt{home}                     & Return to the home screen \\
Back       & \texttt{back}                     & Go back to the previous page \\
Recent     & \texttt{recent}                   & Open recent apps \\
\bottomrule
\end{tabular}
\end{table}

\paragraph{Android-Control-v2.}

AndroidControl-v2 is an offline evaluation benchmark improved upon the original AndroidControl dataset.
It pairs step-by-step task instructions, mobile interface screenshots, and human demonstration actions,
and is designed to evaluate the step-level action prediction capability of GUI agents in Android environments.
To improve data quality, AndroidControl-v2 filters samples by checking the consistency between annotated actions
and oracle low-level instructions, resulting in 398 cleaned samples from the original 500 samples.

The dataset adopts a unified action space consisting of thirteen action types:
Answer, Click, Select, LongPress, Write, KeyboardPress, Scroll, Swipe, Wait,
NavigateHome, NavigateBack, OpenApp, and Terminate, as shown in \Cref{tab:androidcontrol_v2_action_space}.
These actions cover the fundamental operations required for mobile GUI interaction,
including clicking, long pressing, text input, swiping, scrolling, system navigation,
application launching, and task termination.

\begin{table}[t]
\centering
\caption{Action Space of the AndroidControl-v2 Dataset}
\label{tab:androidcontrol_v2_action_space}
\small
\renewcommand{\arraystretch}{1.15}
\setlength{\tabcolsep}{5pt}
\begin{tabular}{llp{0.43\linewidth}}
\toprule
\textbf{Action} & \textbf{Action Format} & \textbf{Functionality} \\
\midrule
Answer 
& \texttt{answer [text]} 
& Return the final answer to the user query \\

Click 
& \texttt{click [target, x,y]} 
& Tap or click a specific UI element \\

Select 
& \texttt{select [target, option]} 
& Select an option from a list or dropdown menu \\

LongPress 
& \texttt{long\_press [target, x,y]} 
& Press and hold a UI element \\

Write 
& \texttt{write [text, x,y]} 
& Enter text into an input field \\

KeyboardPress 
& \texttt{keyboard\_press [key]} 
& Press a specific keyboard key \\

Scroll 
& \texttt{scroll [direction]} 
& Scroll a view or container \\

Swipe 
& \texttt{swipe [direction, x,y]} 
& Perform a touchscreen swipe gesture \\

Wait 
& \texttt{wait [seconds]} 
& Pause execution for UI updates \\

NavigateHome 
& \texttt{home} 
& Navigate to the device home screen \\

NavigateBack 
& \texttt{back} 
& Press the system back button \\

OpenApp 
& \texttt{open\_app [app]} 
& Launch a specified application \\

Terminate 
& \texttt{terminate [message]} 
& Signal the end of the current task \\
\bottomrule
\end{tabular}
\end{table}

\section{Case Study}\label{app:case}

\begin{figure}[!h]
    \centering
    \includegraphics[width=1\linewidth]{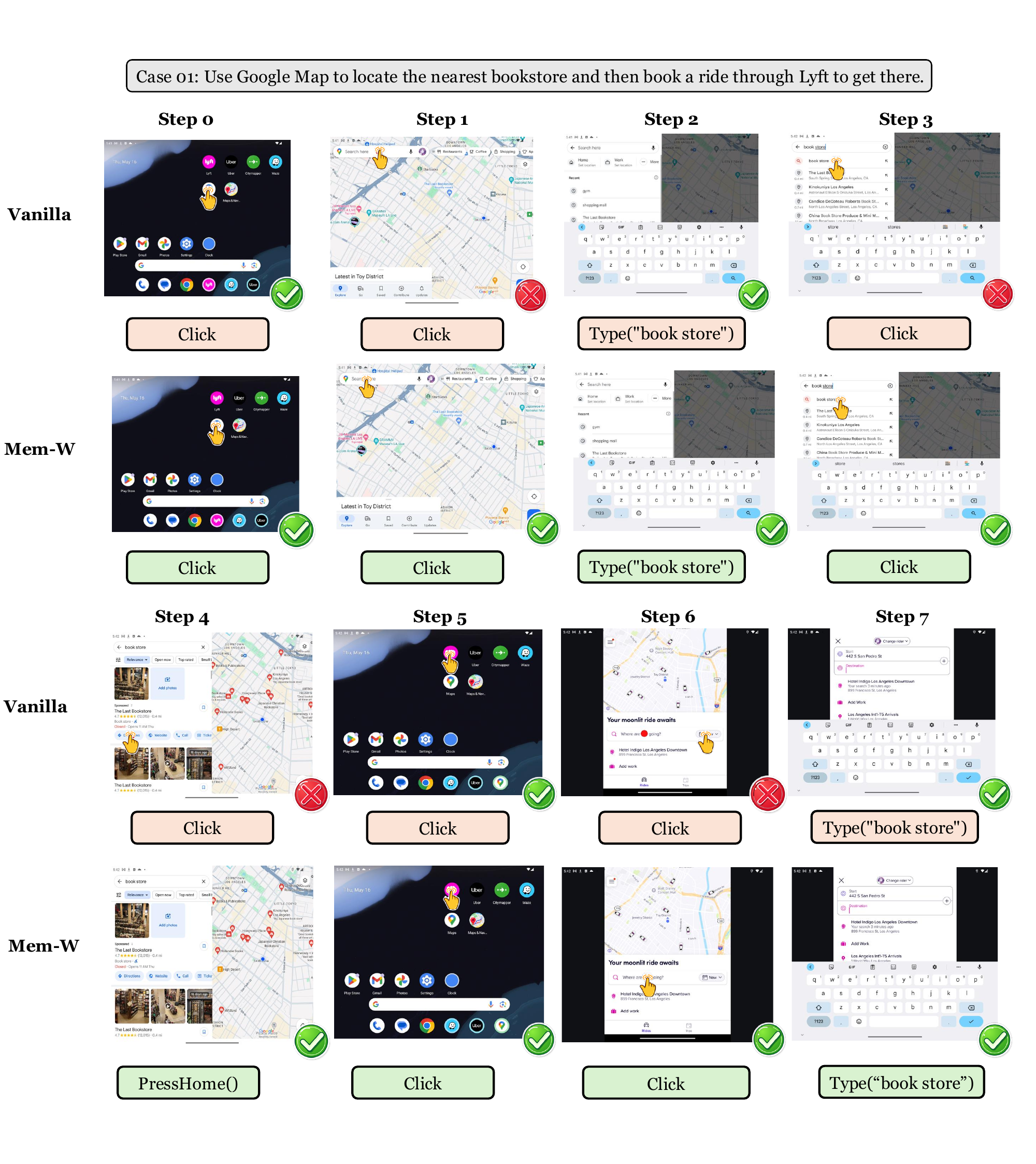}
    \vspace{-1em}
    \caption{Case visualization I (website).}
    \vspace{-1.5em}
    \label{fig:case1}
\end{figure}

\begin{figure}[!h]
    \centering
    \includegraphics[width=1\linewidth]{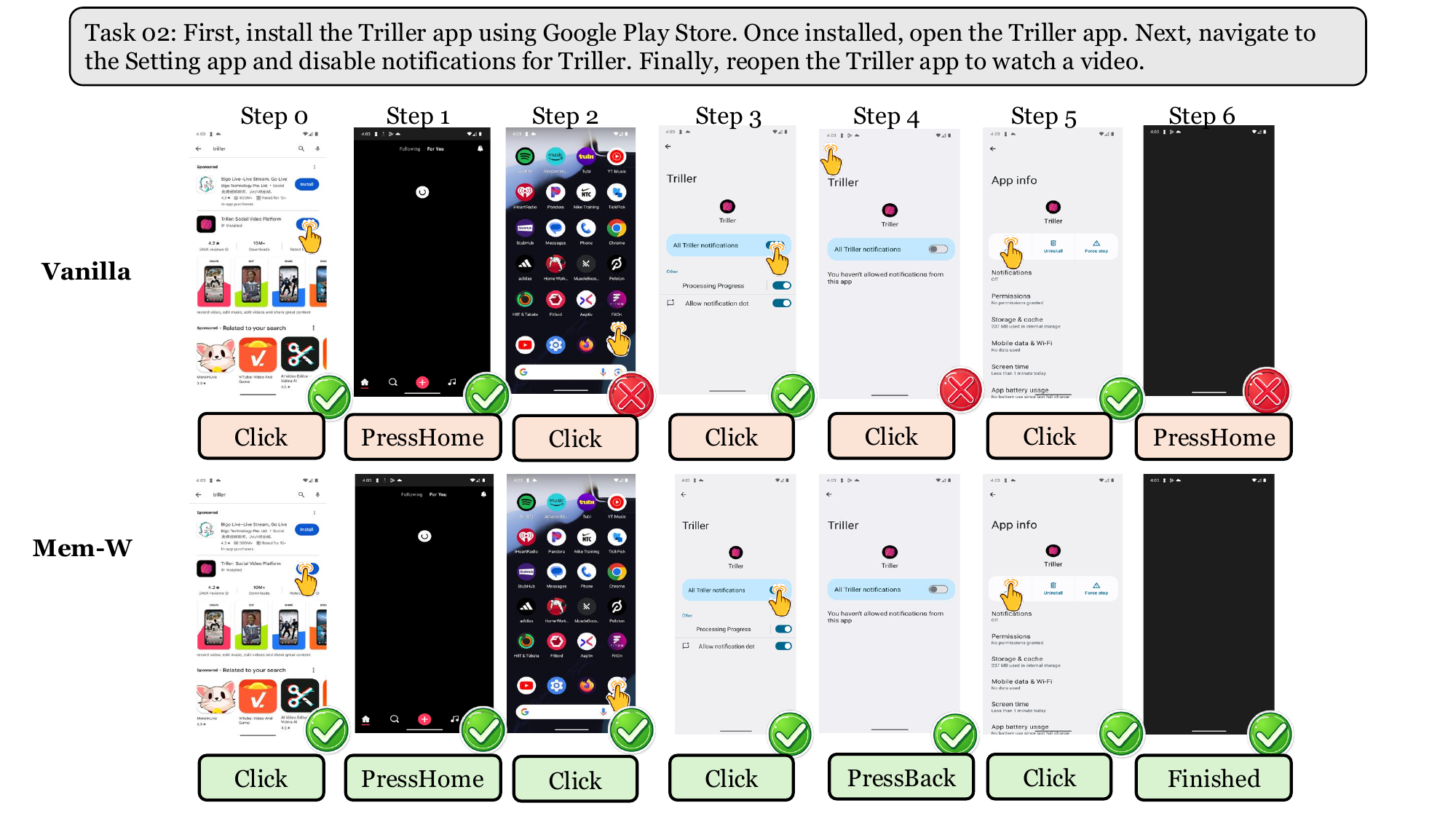}
    \vspace{-1em}
    \caption{Case visualization II (mobile).}
    \vspace{-1.5em}
    \label{fig:case2}
\end{figure}

\begin{figure}[!h]
    \centering
    \includegraphics[width=1\linewidth]{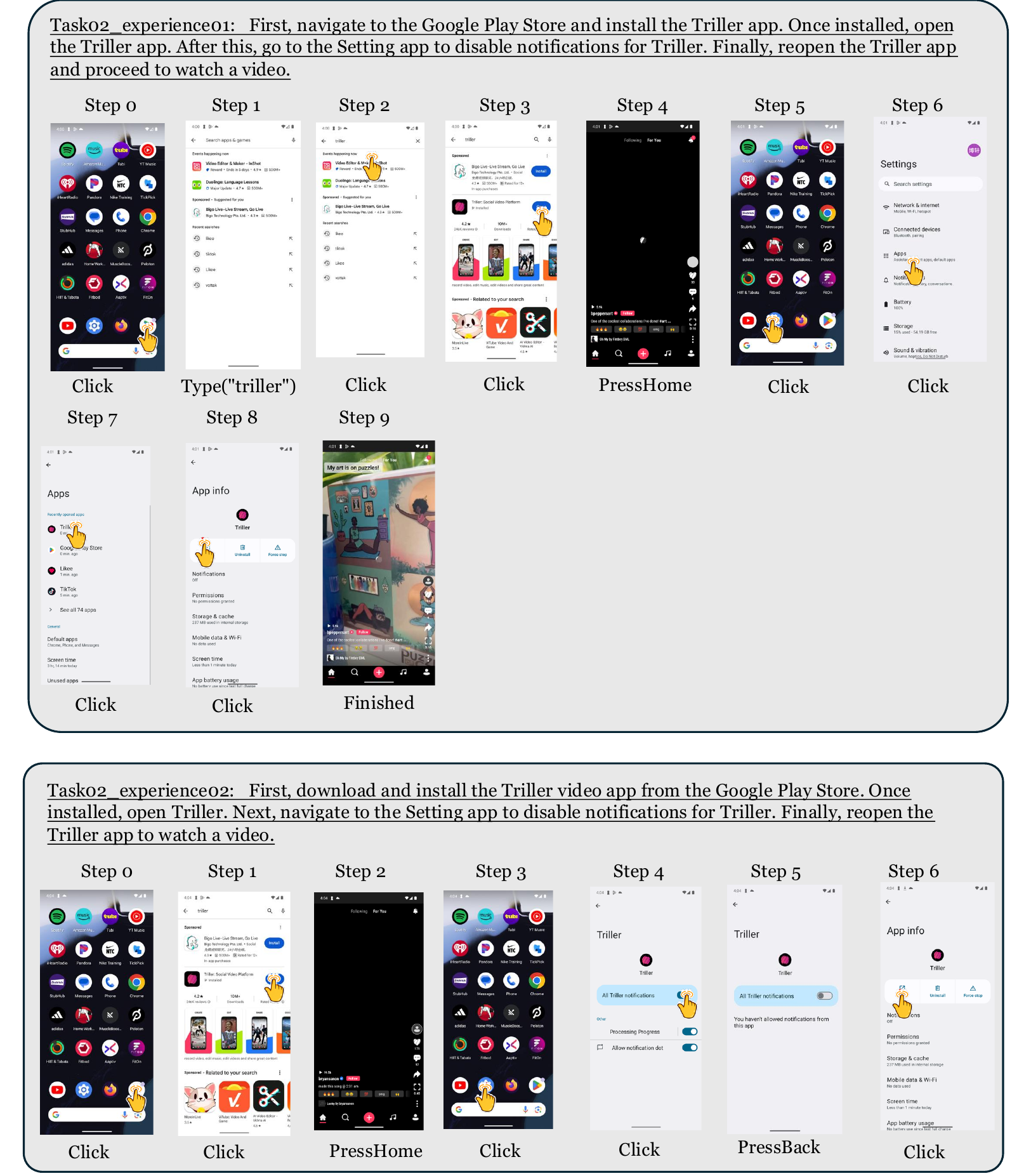}
    \vspace{-1em}
    \caption{Case visualization II (the retrieved trajectories by \ourmethod).}
    \vspace{-1.5em}
    \label{fig:case2-exp}
\end{figure}

We provide qualitative case studies to illustrate how \ourmethod uses latent memory during long-horizon GUI interaction.
\Cref{fig:case1} visualizes a web navigation example, where the agent must preserve task constraints and intermediate progress across multiple page transitions.
\Cref{fig:case2} presents a mobile interaction example, showing how the agent maintains an evolving record of the current episode while selecting executable GUI actions.
To further reveal the role of experiential memory, \Cref{fig:case2-exp} shows the trajectories retrieved by \ourmethod for the mobile case; these retrieved examples provide procedural evidence that complements the in-session working memory and guides the policy toward more reliable action choices.

\section{Limitation}\label{app:limitation}
While \ourmethod demonstrates consistent improvements across diverse benchmarks, several aspects remain open for future investigation. First, the current trajectory-to-latent compressor is trained with a fixed latent-token budget, and adaptively allocating compression capacity based on trajectory complexity could further improve information retention. Also, although we evaluate on four benchmarks spanning web and mobile environments, the generalization of the learned latent memory to other GUI modalities such as desktop applications can also be explored, which we leave for future work. Finally, we view latent memory as a potentially model-agnostic interface rather than a mechanism tied to a particular backbone or tokenizer. Future work could explore more tokenizer-independent latent-token representations that can be transferred across models.
\end{document}